\documentclass[11pt]{article}

\usepackage[final]{acl}

\usepackage{times}
\usepackage{latexsym}
\usepackage[T1]{fontenc}
\usepackage[utf8]{inputenc}
\usepackage{microtype}
\usepackage{inconsolata}
\usepackage{graphicx}
\usepackage{amsmath}
\usepackage{amsfonts}
\usepackage{amssymb}  
\usepackage{stmaryrd} 
\usepackage{booktabs}
\usepackage{multirow}
\usepackage{url}
\usepackage{xcolor}
\usepackage[most]{tcolorbox}
\usepackage{fvextra}
\usepackage{amsthm}

\newtheorem{theorem}{Theorem}
\newtheorem*{theorem*}{Theorem}

\DefineVerbatimEnvironment{promptblock}{Verbatim}{%
  fontsize=\small,
  breaklines=true,
  breakanywhere=true
}

\newcommand{\ours}{\textsc{AlgoGen}}
\newcommand{\vta}{\textsc{VTA}}           
\newcommand{\vtajson}{\textsc{VTA}-JSON}  
\newcommand{\rsl}{\textsc{RSL}}
\newcommand{\projecturl}{\url{https://algenlab.github.io/algogen/}}

\title{ALGOGEN: Tool-Generated Verifiable Traces for Reliable Algorithm Visualization}

\author{
Kunpeng Liao$^{1,\dagger}$ \quad
Yuexiao Ma$^{1,\dagger}$ \quad
Yisheng Lin$^{1}$ \quad
Hualin Zeng$^{1}$ \quad
Xiawu Zheng$^{1}$ \quad
Rongrong Ji$^{1,*}$ \\
$^{1}$Key Laboratory of Multimedia Trusted Perception and Efficient Computing,\\
Ministry of Education of China, Xiamen University, 361005, P.R. China 
}

\begin{document}
\maketitle

\begin{abstract}

Algorithm Visualization (AV) helps students build mental models by animating algorithm execution states. Recent LLM-based systems such as CODE2VIDEO generate AV videos in an end-to-end manner. However, this paradigm requires the system to simultaneously simulate algorithm flow and satisfy video rendering constraints (element layout, color schemes, etc.), a complex task that induces LLM hallucinations. This results in reduced execution success rates, element overlap, and inter-frame inconsistencies.
To address these challenges, we propose ALGOGEN, a novel paradigm that decouples algorithm execution from rendering. We first introduce Visualization Trace Algebra (VTA), a monoid over algorithm visual states and operations. The LLM then generates a Python tracker that simulates algorithm flow and outputs VTA-JSON traces, a JSON encoding of VTA. For rendering, we define a Rendering Style Language (RSL) to templatize algorithm layouts. A deterministic renderer then compiles algorithm traces with RSL into Manim, LaTeX/TikZ, or Three.js outputs\footnote{
Manim, TikZ, and Three.js are respectively a Python animation engine, a LaTeX vector graphics package, and a JavaScript 3D rendering library.}.
Evaluated on a LeetCode AV benchmark of 200 tasks, ALGOGEN achieves an average success rate improvement of $17.3\%$ compared to end-to-end methods ($99.8\%$ vs. $82.5\%$). These results demonstrate that our decoupling paradigm effectively mitigates LLM hallucinations in complex AV tasks, providing a more reliable solution for automated generation of high-quality algorithm visualizations. Demo videos and code are available at: \projecturl.

\end{abstract}

\begin{figure*}[t]
  \centering
  \includegraphics[width=0.9\textwidth]{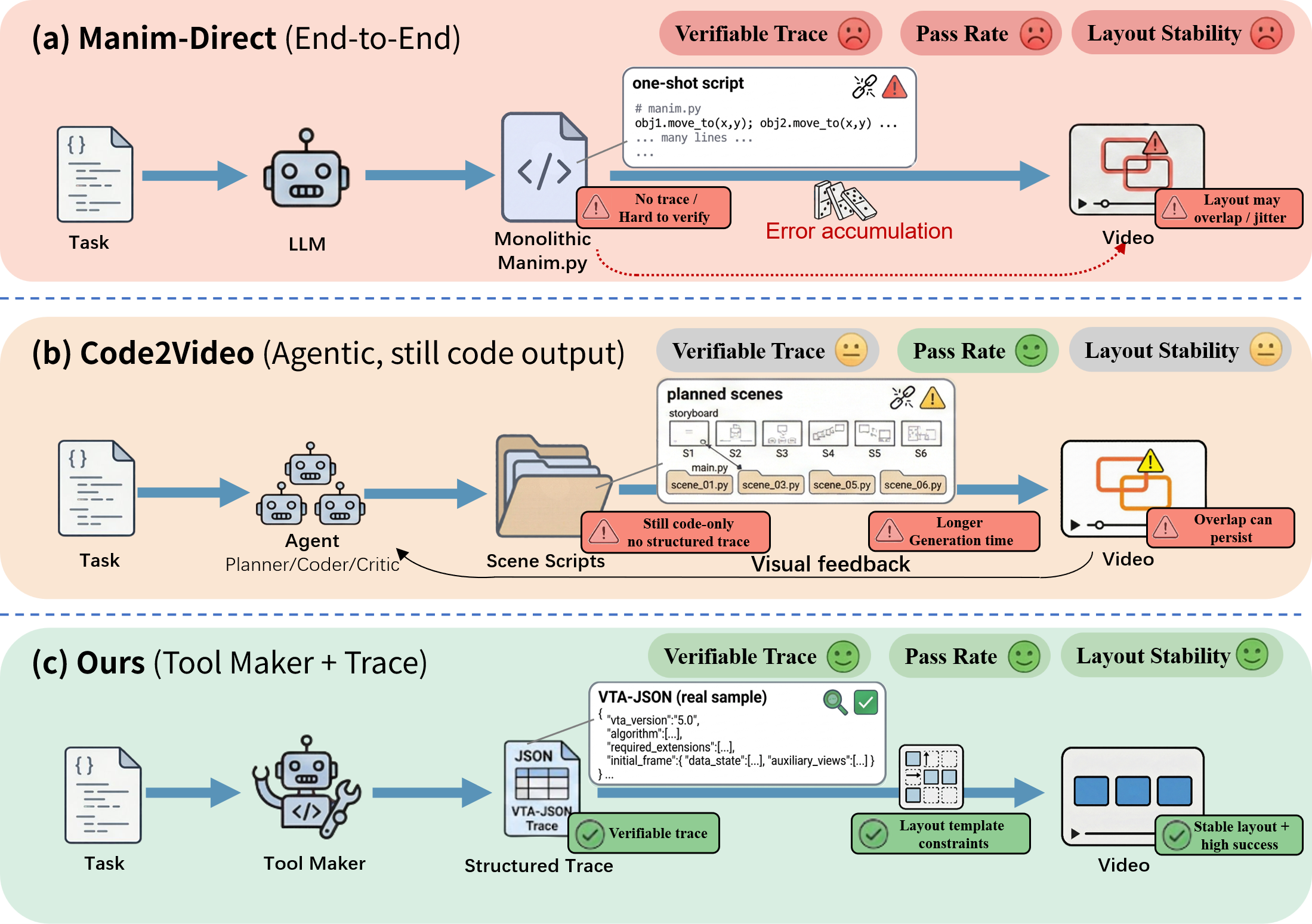}
  \caption{Paradigm comparison. (a) Manim-Direct outputs a monolithic Manim script (no traces $\rightarrow$ error accumulation and layout jitter/overlap). (b) Code2Video plans multiple \texttt{scene\_*.py} files with visual feedback, but still code-only (no schema-validated trace) and slower. (c) Ours outputs an executable tracker + schema-validated \vta{}-JSON~5.0 trace, rendered under \rsl{} templates for stable, verifiable videos.}
  \label{fig:paradigm-shift}
\end{figure*}

\section{Introduction}
Algorithm visualization (AV)~\cite{shaffer2010algorithm,naps2000jhave} helps learners bridge the gap between abstract pseudocode and concrete mental models by animating execution states, with empirical studies showing improved learning outcomes~\cite{naps2002exploring,HUNDHAUSEN2002259}. Classical systems such as JHAVE~\cite{naps2000jhave} and VisuAlgo~\cite{halim2012_unifiedvis} provide textbook-style animations for sorting, graphs, and dynamic programming. However, creating high-quality AV remains costly: instructors must design example inputs, select states to highlight, implement visualization logic in specific toolkits (e.g., Manim~\cite{manim_community}), and iterate on layout and aesthetics. This cost prevents scaling AV coverage to thousands of algorithm problems on online judges and MOOCs~\cite{shaffer2010algorithm,halim2012_unifiedvis}, even as students continue reporting difficulties with core concepts~\cite{10.1145/3230977.3231005}. Meanwhile, LLMs have demonstrated strong code generation capabilities~\cite{chen2021evaluatinglargelanguagemodels} and now power educational tools like GPTutor~\cite{chen2023gptutorchatgptpoweredprogrammingtool}, raising a natural question: can LLMs automatically generate high-quality algorithm visualizations from problem descriptions at scale?
While LLMs show promise in generating charts~\cite{dibia-2023-lida}, algorithm visualization presents unique challenges due to complex state dynamics, making this question both non-trivial and potentially transformative for democratizing algorithm teaching materials.

Recent systems such as TheoremExplainAgent~\cite{ku2025theoremexplainagentvideobasedmultimodalexplanations} and CODE$2$VIDEO~\cite{chen2025code2videocodecentricparadigmeducational} prompt LLMs to produce AV content end-to-end: given an algorithm, the model directly outputs Manim code that is rendered into animation.
However, this approach suffers from four limitations:
\textbf{($1$) Unverifiable correctness}---the pipeline produces only code-only Manim scripts (monolithic or multi-scene) without structured traces of intermediate states, making systematic verification difficult and forcing manual spot-checking;
\textbf{($2$) Layout instability}---LLMs place visual elements with ad-hoc coordinates, causing overlaps and frame-to-frame jitter due to the lack of consistent canvas-level planning;
\textbf{($3$) Low reuse}---fixing errors requires regenerating entire scripts without reusable intermediate representations;
and \textbf{($4$) Latent execution complexity}---end-to-end generation forces LLMs to implicitly simulate execution without explicit chain-of-thought reasoning~\cite{wei2023chainofthoughtpromptingelicitsreasoning}, exceeding their ability to jointly satisfy algorithmic correctness, rendering API calls, and layout constraints.
On our $200$-task LeetCode benchmark, the Manim-Direct baseline achieves only $82.5\%$ rendering success, suggesting the core bottleneck is not lack of algorithmic knowledge but \emph{task complexity}: end-to-end generation creates a long-horizon planning problem with large action spaces where small early deviations cascade across subsequent steps, preventing consistent generation of correct and readable scripts.

To address these challenges, we decompose AV generation into smaller sub-tasks---tracker generation, execution trace validation, and deterministic rendering---narrowing the action space so each step has clear, standardized outputs.
Drawing on augmented language models~\cite{mialon2023augmentedlanguagemodelssurvey, schick2023toolformerlanguagemodelsteach}, Program-of-Thoughts~\cite{chen2023programthoughtspromptingdisentangling}, and LLMs as tool makers~\cite{cai2024latm,zhang2024agentoptimizer}, we propose the \emph{LLM Tool Generation} paradigm: the LLM acts as a tool maker rather than direct video generator.
First, the LLM generates a Python tracker that emits execution information in \vtajson{} $5.0$, the JSON encoding of our Visualization Trace Algebra (\vta{})---a monoid describing algorithmic visual states and composable operations with schema-validated correctness.
Second, the model generates a Rendering Style Language (\rsl{}) specification that selects safe layout templates and aesthetic styles.
Finally, deterministic renderers interpret \vta{} under \rsl{} guidance to produce Manim videos, LaTeX/TikZ figures~\cite{tantau2025tikzpgf}, or interactive Three.js~\cite{threejs2025r182} scenes.

This decomposition offers four advantages:
\textbf{($1$) Verifiability}---the tracker outputs schema-validated JSON traces testable on multiple inputs and comparable to reference implementations, enabling systematic automated correctness checking;
\textbf{($2$) Reusability}---a single tracker serves multiple inputs and rendering backends (Manim, TikZ, Three.js), amortizing LLM cost and ensuring consistent visual design;
\textbf{($3$) Debuggability}---errors localize to specific trace steps with structured diagnostics for targeted retry, avoiding monolithic script editing;
and \textbf{($4$) Separation of concerns}---LLMs focus on algorithmic structure and pedagogical sequencing while layout and aesthetics are delegated to domain-specific algorithms and \rsl{}-guided styling.

\begin{figure*}[t]
  \centering
  \includegraphics[width=1.0\textwidth]{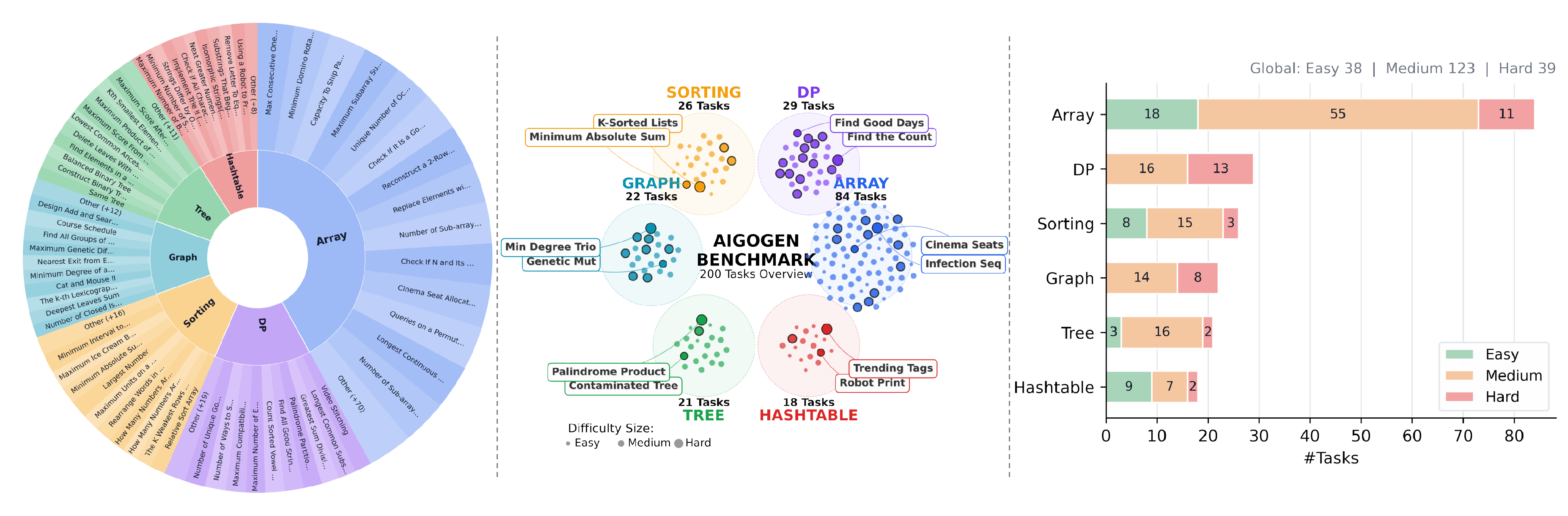}
  \caption{\textbf{Overview of ALGOGEN-Bench.} (1) \textbf{Left:} taxonomy of 200 LeetCode tasks across six algorithm families. (2) \textbf{Middle:} dataset coverage map; nodes denote tasks colored by family and sized by difficulty (Easy/Medium/Hard). (3) \textbf{Right:} difficulty breakdown per family (Global: 38 Easy, 123 Medium, 39 Hard).}
  \label{fig:dataset_overview}
\end{figure*}

To rigorously evaluate our framework, we introduce \ours{}-Bench, a standardized benchmark derived from a large corpus of LeetCode problems collected via LeetCode's GraphQL endpoint. We convert problems into our unified task format and stratify-sample $200$ tasks spanning six major algorithm families (Array, DP, Sorting, Graph, Tree, Hashtable) and multiple difficulty levels, with manual relabeling and test-case verification for reliability. Figure~\ref{fig:dataset_overview} summarizes the benchmark taxonomy, coverage, and difficulty composition. Overall, \ours{}-Bench enables reproducible evaluation of schema-validated, step-wise execution traces.

\begin{figure*}[t]
  \centering
  \includegraphics[width=0.9\textwidth]{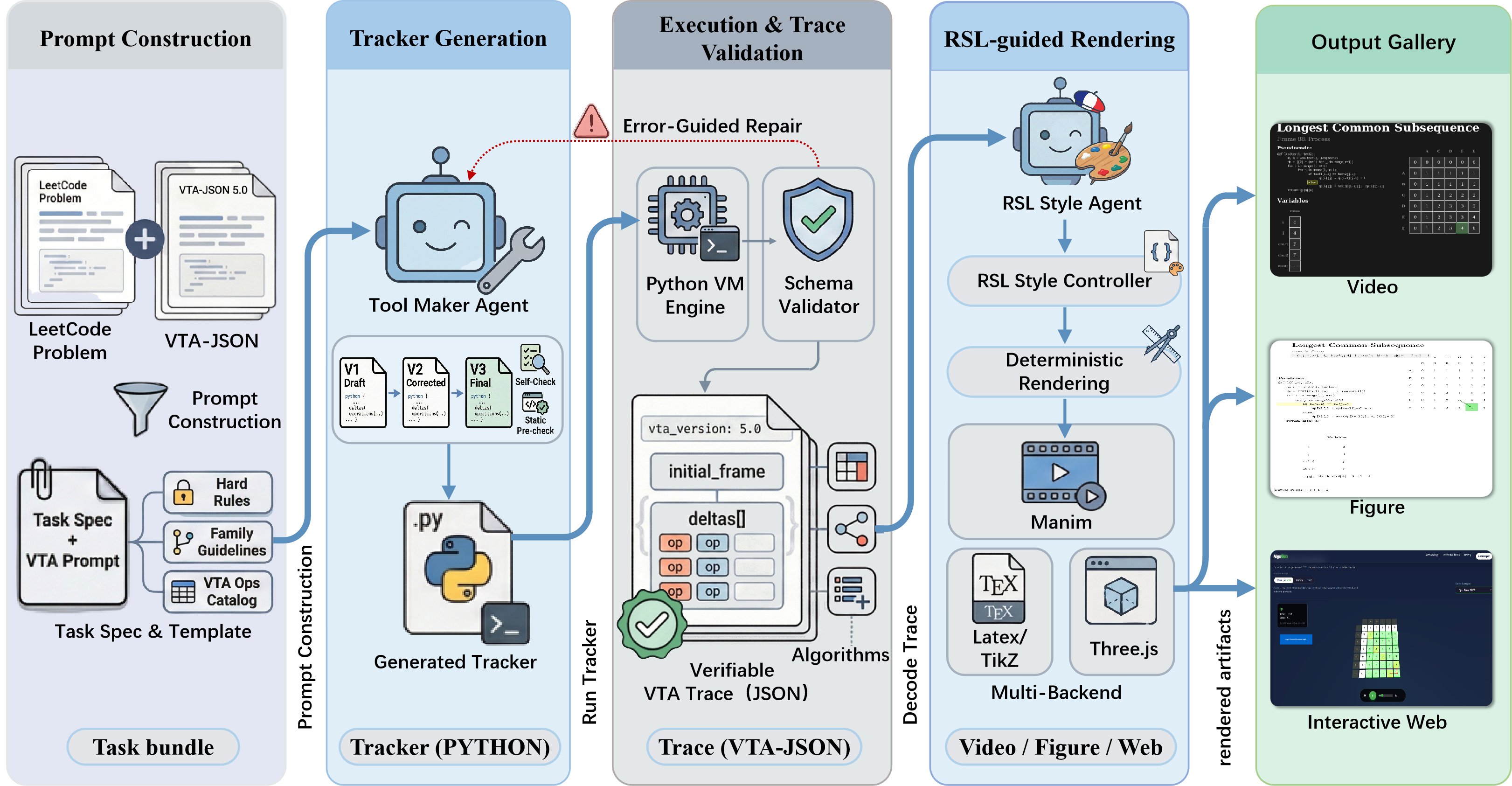}
  \caption{Overview of \ours{}: Prompt Construction $\rightarrow$ Tracker Generation $\rightarrow$ Execution \& Trace Validation $\rightarrow$ \rsl{}-guided Rendering. Validation failures trigger error-guided repair; \rsl{} controls style/layout without changing trace semantics; deterministic backends include Manim, LaTeX/TikZ, and Three.js.}
  \label{fig:pipeline}
\end{figure*}

Our contributions are as follows:

\begin{enumerate}
  \item We introduce a framework that decomposes AV generation into tracker generation, trace validation, and deterministic rendering, enabling verifiable correctness, cross-backend reusability, and explainable debugging.
  
  \item We formalize a monoid-based algebra for algorithmic visual states and composable operations, encoded as schema-validated \vtajson{}~$5.0$, which decouples algorithmic semantics from rendering backends and enables automated correctness checking.
  
  \item We design a Rendering Style Language that automates layout planning and aesthetic refinement, eliminating ad-hoc coordinates and ensuring collision-free compositions across Manim, LaTeX/TikZ, and Three.js.
  
  \item We construct \ours{}-Bench with $200$ diverse LeetCode problems, demonstrating $99.8\%$ pipeline success and $99.2\%$ algorithmic correctness vs.\ $82.5\%$ and $87.0\%$ for end-to-end baselines, establishing state-of-the-art for LLM-driven algorithm visualization.
\end{enumerate}

\section{Related Work}
\label{sec:related}

\textbf{Classical Algorithm Visualization.}
Classical AV systems and program tracers like Python Tutor~\cite{40591} demonstrate pedagogical benefits~\cite{naps2002exploring}.
Subsequent work has proposed large web-based platforms such as VisuAlgo and related unified environments for teaching data structures and algorithms~\cite{halim2012_unifiedvis}, as well as dynamic, activity-focused AV systems and game-based visualizations~\cite{VRACHNOS2014229,10.1145/3408877.3432520}.
However, these systems require substantial manual effort per algorithm, and coverage lags far behind the thousands of exercises on modern online judges~\cite{10.1145/3230977.3231005}. Recent platforms also integrate conversational agents with dynamic algorithm visualizations (e.g., VisualCodeMOOC)~\cite{li2025visualcodemooc}.

\textbf{LLM-Based Educational Content.}
LLMs have been applied to generate explanations and visualizations.
TheoremExplainAgent (TEA) generates Manim videos for math proofs, and CODE$2$VIDEO explains code concepts via automatically generated animations~\cite{ku2025theoremexplainagentvideobasedmultimodalexplanations,chen2025code2videocodecentricparadigmeducational}.
Beyond video generation, LLMs have also been deployed as interactive programming tutors and teaching assistants, for example GPTutor for line-by-line code explanation in VS~Code~\cite{chen2023gptutorchatgptpoweredprogrammingtool} and studies on using ChatGPT as a teaching assistant in data-structures-and-algorithms courses~\cite{Jamie_2025}.
Unlike \ours{}, which uses a structured IR (\vta{}) to capture runtime state dynamics and ensures verifiability, these systems rely on end-to-end generation (TEA) or AST-based templates (CODE$2$VIDEO), limiting their ability to handle complex algorithm execution reliably.

\textbf{LLM Code Generation.}
While LLMs excel at code generation, they often struggle with complex APIs (e.g., Manim) due to hallucinations~\cite{zhong2024chatgptreplacestackoverflowstudy,liu2023codegeneratedchatgptreally}.
Recent reviews and surveys highlight both rapid progress and brittleness when code LLMs are used in real-world development workflows~\cite{HUSEIN2025103917,10.1145/3747588,weber2024largelanguagemodelssoftware}.
Although retrieval-based methods such as DocPrompting~\cite{zhou2023docpromptinggeneratingcoderetrieving} help with library use, our analysis shows that adding RAG documentation often fails to fix API misuse \emph{in practice}, especially under complex visual logic.
More broadly, tool-using LLM agent frameworks improve reliability by delegating execution to external tools~\cite{ge2023openagillmmeetsdomain,liu2024dynamicllmpoweredagentnetwork,zhang2024agentoptimizer}, while recent work also studies how to diagnose failures in LLM multi-agent systems~\cite{zhang2025agentfailure}.
LLM-based data visualization similarly combines IRs with deterministic rendering~\cite{ouyang2025nvagentautomateddatavisualization,zhao2025chartcoderadvancingmultimodallarge}.
Beyond our immediate task setting, recent AI systems have also explored reliability--efficiency trade-offs in model compression and search, including LLM quantization, mixed-precision quantization, ViT post-training quantization, pruning, and training-free transformer architecture search~\cite{ma2024affinequant,ma2023ompq,ma2023oscillation,ma2024outlieraware,zheng2025itpruner,zhou2021ecdarts,zhou2022tftas,zhou2024trazor,zhong2024isvit,zhong2025erq}.
Related trends also appear in efficient generative modeling and synthetic-content analysis, such as autoregressive video acceleration, test-time diffusion error correction, AI-generated text detection, visual-linguistic face forgery detection, and multimodal retrieval/re-identification~\cite{ma2026flowcache,zhong2026iec,sun2026detectrouter,sun2026latestage,sun2025vlffd,sun2025continualffd,tan2024rle,tan2025aga,feng2025mdreid,li2025mfrnet,li2026find,sun2024diffusionfake}.
Instead, we have the LLM generate simple \emph{tracker} code and shift complexity to deterministic renderers.

\section{Method}
\label{sec:method}

\subsection{Overview}
\label{subsec:overview}

Given an algorithm visualization task $T$ (an algorithm problem description with auxiliary metadata), our goal is to produce a visualization artifact $V$ (video/figure/interactive page). As shown in Figure~\ref{fig:pipeline}, our key idea is to make the intermediate process controllable and verifiable: instead of synthesizing renderer-level animations end-to-end, we (i) represent executions as schema-validated \vtajson{} traces (\vta{}; Sec.~\ref{subsec:vta}), (ii) prompt an LLM to generate an executable Python tracker that emits these traces during execution (Sec.~\ref{subsec:trackers}), and (iii) render validated traces with deterministic backends under an optional style specification in \rsl{} (Sec.~\ref{subsec:rsl}). This decomposition separates \emph{what happens} (execution semantics) from \emph{how it looks} (presentation), enabling validation and localized repair while keeping rendering deterministic.

\subsection{\vta{}: Visualization Trace Algebra}
\label{subsec:vta}

\paragraph{Motivation and usage (I/O).}
We introduce \vta{}, a schema-validated \vtajson{} trace IR, because end-to-end renderer-level animation generation entangles long-horizon algorithm semantics with low-level rendering and layout decisions, making errors cascade across steps; \vta{} instead records execution as locally checkable, typed transitions, enabling modular verification/repair and deterministic replay across backends, and empirically improves reliability over end-to-end rendering in our experiments.
Unlike declarative grammars for static visualization (e.g., Vega-Lite~\cite{7539624}), \vta{} directly models step-wise mutations (e.g., pointer movements and updates to auxiliary structures).
In our pipeline, a Python tracker executes the algorithm and emits a trace: \textbf{input} is the tracker’s runtime algorithm state (e.g., arrays/graphs/DP tables and auxiliary variables) together with the current pseudocode line (for highlighting), and \textbf{output} is a schema-validated \vtajson{} file (\texttt{trace.json}) that deterministic renderers can consume to produce the final artifact.

\paragraph{Trace format and validation contract.}
An algorithm run is represented as a trace $(s_0, w)$ and serialized in \vtajson{}: \texttt{initial\_frame} encodes the initial visual state $s_0$, and \texttt{deltas} records the evolution as a sequence of small, typed operation batches.
Each \texttt{delta} is aligned with a highlighted pseudocode line, and concatenating all deltas yields the full operation sequence $w$.
We store traces as deltas (rather than absolute frames), aligning naturally with algorithm steps and animation primitives.
A validator enforces \vtajson{} invariants---including version consistency, referential integrity, and type-correct operation parameters~\cite{lu2025learninggeneratestructuredoutput}---and traces must pass validation before deterministic rendering; otherwise, we trigger targeted error-guided repair using validator diagnostics.
The state space is many-sorted (arrays, graphs, trees, hash tables, DP tables, and auxiliary views), and \vta{} provides a compact operation catalogue covering style updates, structural updates, and explanatory overlays (Appendix~\ref{sec:appendix-vtajson}).

\paragraph{Theoretical analysis.}
We model an algorithm visualization as a typed visual state $s \in S$ acted on by primitive operations.
Each primitive operation symbol $o \in Op$ denotes a (partial) state transformer $\llbracket o \rrbracket : S \rightharpoonup S$.
A finite sequence of operations $w = o_1\ldots o_n \in Op^*$ acts on $s$ by composition:
\[
s \cdot \epsilon = s,\qquad
s \cdot (o_1 \ldots o_n) =
\llbracket o_n \rrbracket \circ \dots \circ \llbracket o_1 \rrbracket (s),
\]
where $\epsilon$ is the empty sequence.
This view matches our trace format: each \texttt{delta} encodes a small word in $Op^*$, and a full \vtajson{} trace is obtained by concatenating these words.
Deterministic renderers then interpret the trace by sequentially applying the corresponding state transformers.

\begin{theorem}
Let $Op$ be the primitive operation set in our \vtajson{} specification, and let $*$ be concatenation on $Op^*$ with identity $\epsilon$.
Then $(Op^*, *, \epsilon)$ satisfies the monoid axioms:
\[
\begin{aligned}
&\forall u,v,w\in Op^*, &&(u*v)*w = u*(v*w), \\
&\forall u\in Op^*, &&\epsilon*u = u, \quad u*\epsilon = u.
\end{aligned}
\]
\label{the:monoid}
\end{theorem}
\noindent Where $Op^*$ is the set of all finite sequences over $Op$ (including $\epsilon$). Proof of Theorem~\ref{the:monoid} in Appendix~\ref{sec:appendix-vta}. The monoid structure justifies treating delta batches as composable trace fragments: concatenation yields a well-defined trace independent of parenthesization.
As a result, our pipeline can validate, debug, and render traces modularly while preserving execution semantics.

\subsection{Tool Maker: LLM-Generated Trackers}
\label{subsec:trackers}

\noindent\textbf{Tracker Synthesizer.}
Because end-to-end renderer-level generation hides intermediate execution state and entangles algorithm semantics with low-level API/layout decisions, we instead use the LLM to generate a runnable tracker that executes the algorithm and emits a schema-validated trace.
This module prompts the LLM to synthesize an instrumented Python tracker that emits \vtajson{} deltas, with execution/validation feedback enabling error-guided repair.
\textbf{Input} is the Unified Task Bundle (task specification, test input, pseudocode, and the required \vtajson{} schema), and \textbf{output} is an executable \texttt{tracker.py} that produces a \vtajson{} trace (\texttt{trace.json}) when run.
The tracker executes the algorithm on the task's input and emits \vtajson{} deltas via a lightweight \texttt{Visualizer} wrapper.
The tracker is responsible for \emph{what happens} during execution (i.e., correct state transitions and aligned highlights), while leaving \emph{how it looks} to deterministic renderers.

To improve robustness, we use a three-stage generation strategy: (i) draft, (ii) self-refinement, and (iii) error-guided repair driven by execution errors and schema-validation feedback~\cite{madaan2023selfrefineiterativerefinementselffeedback}. We also apply a lightweight static pre-check before execution. Full prompt templates and implementation details are provided in Appendix~\ref{sec:appendix-prompts}~\cite{wang2023selfconsistencyimproveschainthought}.

\subsection{Style Controller: \rsl{}-Guided Deterministic Rendering}
\label{subsec:rsl}

\noindent\textbf{Style Controller.}
To avoid the layout instability and error accumulation of end-to-end script synthesis, we restrict the model to producing a high-level, schema-based style specification (\rsl{}), while deterministic backends handle low-level graphics decisions when mapping validated traces to artifacts.
Style Controller uses an LLM to specify high-level layout and aesthetics (\rsl{}), while deterministic renderers map the validated trace to videos/figures/pages. \rsl{} is a compact JSON-based domain-specific language (DSL) that captures high-level layout and style preferences (e.g., \texttt{layout}, \texttt{theme}, \texttt{annotations}) and is generated once per task as a JSON script. Given lightweight trace metadata (e.g., algorithm family and trace length/\#frames), the LLM generates an \rsl{} instance conforming to a fixed schema; deterministic backends then validate and interpret this \rsl{} into renderer-specific configuration, choosing suitable layouts and pacing without changing trace semantics. As a result, the model controls high-level appearance without issuing low-level graphics API calls. The \rsl{} schema is given in Appendix~\ref{sec:appendix-rsl}.

\begin{table}[t]
  \centering
  \small
  \begin{tabular}{lc}
    \toprule
    Method & Algorithm correctness \\
    \midrule
    Ours (VTA+RSL) & $\mathbf{99.8\%}$ \\
    Ours (VTA-only) & $99.2\%$ \\
    \midrule
    \texttt{manim\_direct} & $87.0\%$ \\
    \texttt{manim\_direct\_novta} & $84.9\%$ \\
    \bottomrule
  \end{tabular}
  \caption{LLM-judged algorithm correctness for our systems and end-to-end Manim baselines.}
  \label{tab:tier2-correctness}
\end{table}

\begin{table*}[t]
  \centering
  \small
  \begin{tabular}{lccc}
    \toprule
    System / Model & Metric & Success (\# / $200$) & Videos / Prompt size \\
    \midrule
    DeepSeek-V$3.1$ & Trace success (ours) & $200$ / $200$ (\textbf{$100.0\%$}) & $200$ AES/TEA videos \\
    Qwen$3$-$235$B    & Trace success (ours) & $199$ / $200$ (\textbf{$99.5\%$})  & $199$ AES/TEA videos \\
    GLM-$4.6$       & Trace success (ours) & $200$ / $200$ (\textbf{$100.0\%$}) & $200$ AES/TEA videos \\
    Average       & Trace success (ours) & $599$ / $600$ (\textbf{$99.8\%$})  & $599/600$ AES/TEA videos \\
    \midrule
    Manim-Direct (No RAG)       & Video rendering (e$2$e) & $165$ / $200$ (\textbf{$82.5\%$}) & $165$ videos, $\sim$ $4$k tokens \\
    Manim-Direct (RAG-enhanced) & Video rendering (e$2$e) & $158$ / $200$ ($79.0\%$)           & $158$ videos, $\sim$ $12$k tokens \\
    \bottomrule
  \end{tabular}
  \caption{Success on the 200-task benchmark. Our \vta{} pipeline is evaluated by trace-generation success (valid \vtajson{}), while Manim-Direct baselines are evaluated by end-to-end video rendering success.}
  \label{tab:vta-vs-manim-success}
\end{table*}

\begin{figure}[ht]
  \centering
  \includegraphics[width=\linewidth]{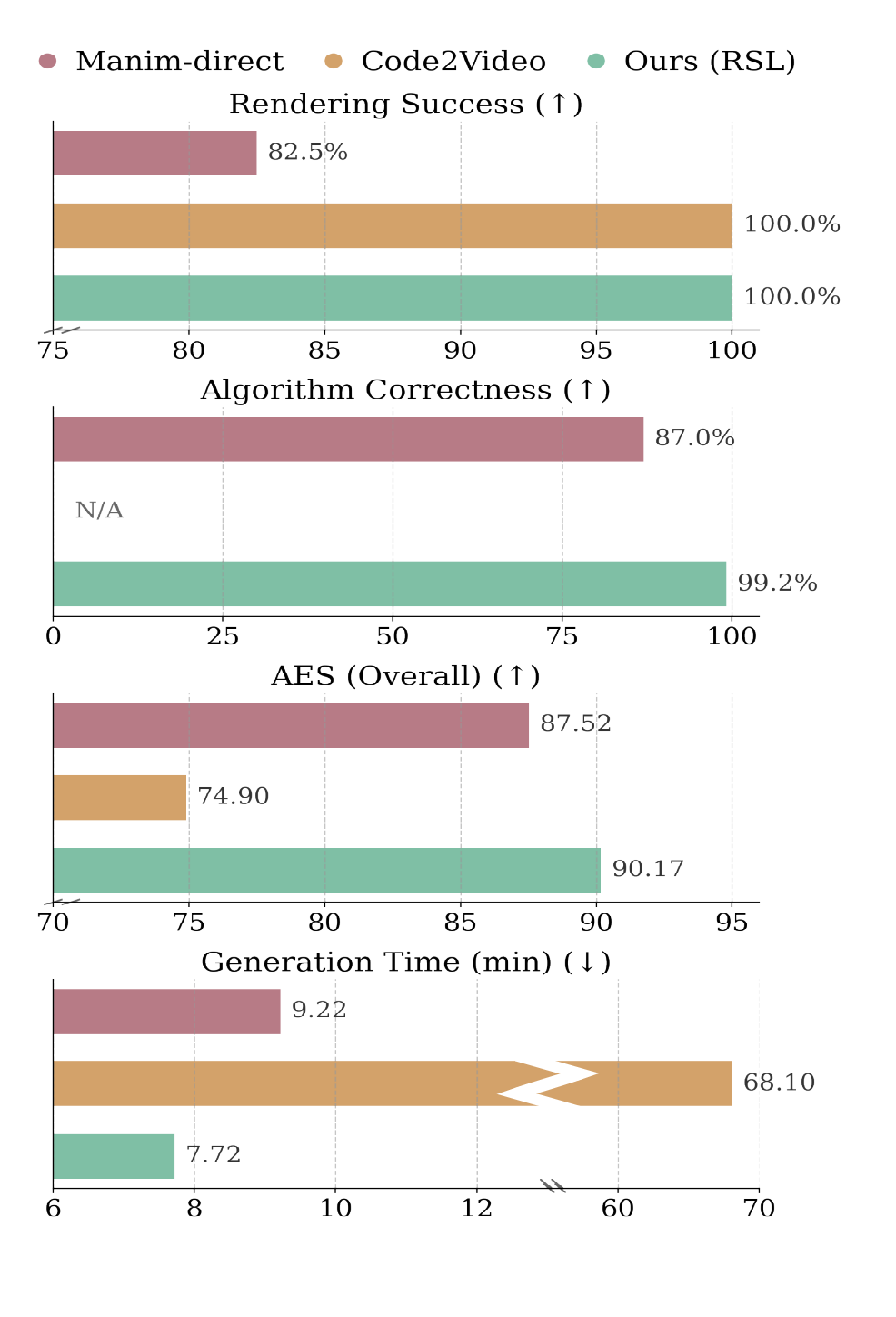}
  \caption{\textbf{System-level comparison across four dimensions.} Rendering success, algorithm correctness, aesthetic quality (AES), and generation time.}
  \label{fig:experiment_summary}
\end{figure}

\section{Experimental Setup}
\label{sec:experiments}

\begin{table*}[t]
  \centering
  \small
  \begin{tabular}{lccc cccccc}
    \toprule
    \multirow{2}{*}{Method} & \multirow{2}{*}{\#Eval} & \multicolumn{2}{c}{Efficiency ($\downarrow$)} & \multicolumn{6}{c}{Aesthetics ($\uparrow$)} \\
    \cmidrule(lr){3-4} \cmidrule(lr){5-10}
     &  & Time (min) & Tokens (K) & EL & AT & LF & VC & AD & Avg. \\
    \midrule
    \textbf{Ours (\vta{} + RSL)} & $200$ & $7.72$ & $31.6$ & $17.70$ & $17.02$ & \textbf{$18.36$} & \textbf{$18.42$} & \textbf{$18.67$} & $\mathbf{90.17}$ \\
    \textbf{Ours (\vta{} Only)} & $599$ & $8.59$ & $20.3$ & $17.36$ & $16.67$ & $18.01$ & $18.03$ & $18.37$ & $88.45$ \\
    manim\_direct & $156$ & $9.22$ & $17.4$ & $17.35$ & $16.48$ & $17.93$ & $17.90$ & $17.87$ & $87.52$ \\
    manim\_direct\_novta & $127$ & $8.91$ & $7.5$ & \textbf{$17.79$} & \textbf{$17.41$} & $18.11$ & $18.28$ & $18.11$ & $89.70$ \\
    Code$2$Video & $200$ & $68.10$ & $24.0$ & $13.75$ & $13.07$ & $15.22$ & $15.41$ & $17.46$ & $74.90$ \\
    \bottomrule
  \end{tabular}
  \caption{AES comparison across systems, including the RSL-enhanced variant. Efficiency: Time (average minutes per task) and Tokens (average token consumption per topic). Aesthetics: Element Layout (EL), Attractiveness (AT), Logic Flow (LF), Visual Consistency (VC), Accuracy \& Depth (AD), and Avg.}
  \label{tab:aes-systems}
\end{table*}

\subsection{Dataset Construction}

We construct a $200$-task LeetCode benchmark with clear algorithmic logic, step-wise visualizable state changes, diverse difficulty (Easy/Medium/Hard), and coverage of six families (sorting, arrays, dynamic programming, trees, graphs, hash tables; Figure~\ref{fig:dataset_overview} shows the taxonomy view, coverage map, and per-family difficulty mix).
Tasks are stored as standardized \texttt{example/*.txt} specifications; Appendix~\ref{sec:appendix-dataset} details the format and quality-control procedures.

\subsection{Models and Baselines}

We use several LLMs as tool makers, including Qwen$3$-$235$B~\cite{yang2025qwen3technicalreport}, DeepSeek V$3.1$~\cite{deepseekai2025deepseekv3technicalreport}, and GLM-$4.6$~\cite{glm2024chatglmfamilylargelanguage}, accessed via the SiliconFlow API~\cite{siliconflow2025api}.
We use the three-stage generation strategy in Section~\ref{sec:method}.
For each task, we generate an initial tracker and allow up to three error-guided repair attempts; the final tracker is reused across different inputs and backends.

We compare our \vta{}-based system with three end-to-end Manim baselines: \textbf{manim\_direct}, where the LLM directly outputs Manim Scene code; \textbf{manim\_direct\_rag}, which adds $12$k-token RAG-enhanced Manim documentation; and \textbf{manim\_direct\_novta}, a less structured baseline without \vta{}-style constraints (only loose prompts).
All baselines share the same maximum tokens, and retry strategy as our method; full prompt templates and implementation details are provided in Appendix~\ref{sec:appendix-baseline-prompts}.

Figure~\ref{fig:qual_merge_stones} presents a qualitative comparison on a representative dynamic-programming task (\emph{Minimum Cost to Merge Stones}).

\begin{figure*}[t]
  \centering
  \includegraphics[width=0.85\textwidth]{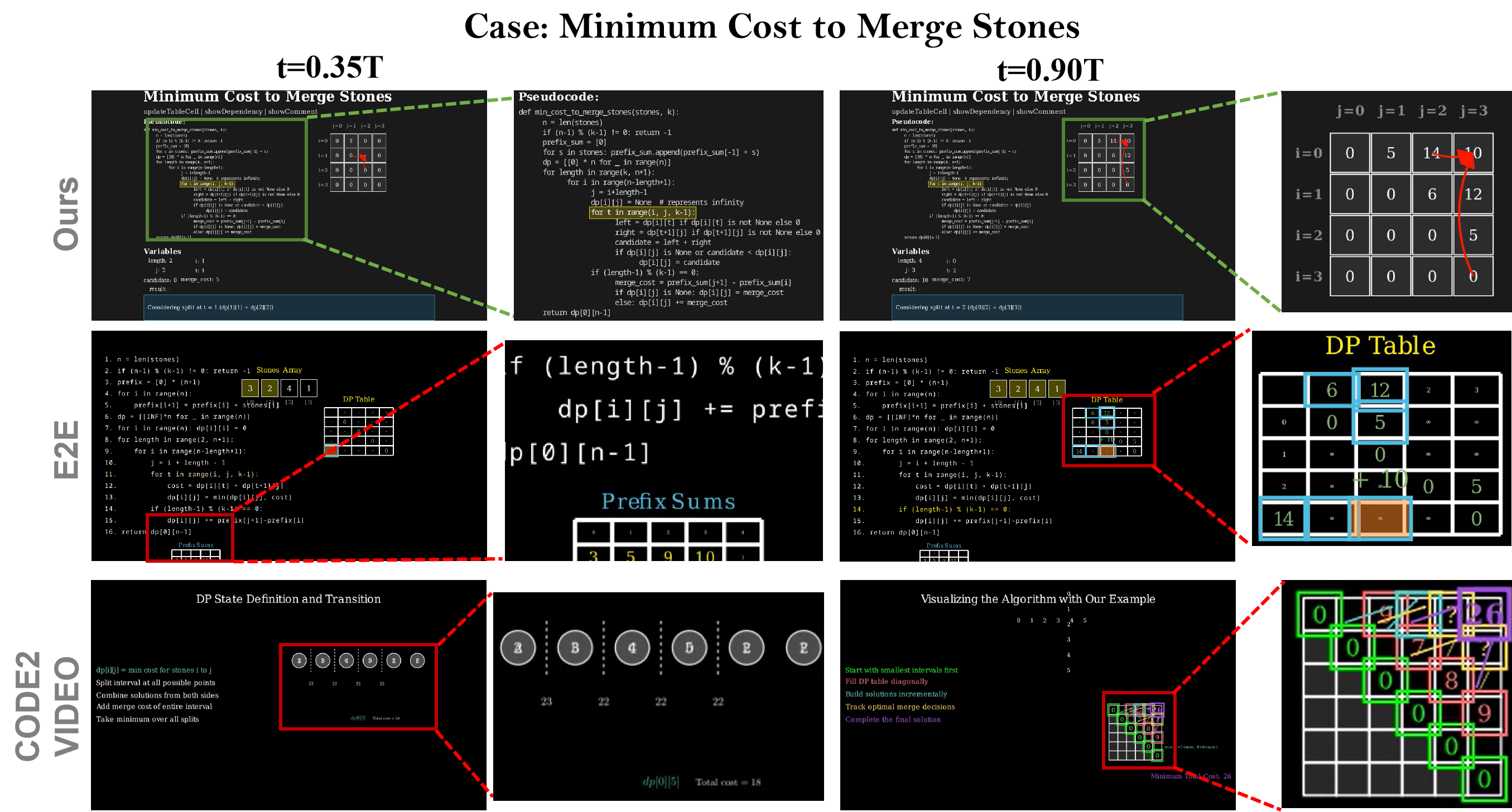}
  \caption{Qualitative comparison on a dynamic-programming task (\emph{Minimum Cost to Merge Stones}).}
  \label{fig:qual_merge_stones}
\end{figure*}

\subsection{Evaluation Metrics}

\paragraph{Rendering success rate.}
The fraction of tasks that complete tracker execution, trace validation, and rendering without unrecoverable exceptions (playable video).

\paragraph{Algorithm correctness.}
Following LLM-as-judge practice~\cite{zheng2023judgingllmasajudgemtbenchchatbot}, a code-level evaluator scores each tracker with a rubric (algorithm logic, \vta{} compliance, presentation); we rescale the $0$--$50$ algorithm-logic score to a $0$--$100$ ``algorithm-correctness'' metric (Table~\ref{tab:tier2-correctness}).

\paragraph{Aesthetic quality (AES).}
We adopt the automatic aesthetic scoring model from CODE$2$VIDEO~\cite{chen2025code2videocodecentricparadigmeducational}, following its AES rubric (5 dimensions) to rate frames and videos on a $0$--$100$ scale.
For each video, we report the average AES over sampled frames.

\paragraph{TEA evaluation.}
We additionally apply the multimodal evaluation rubric from TheoremExplainAgent (TEA)~\cite{ku2025theoremexplainagentvideobasedmultimodalexplanations}, using their $0$--$5$ scale per dimension to assess visual and textual quality in a way comparable to prior work on educational video generation.

\section{Results}
\label{sec:results}

\subsection{Overall Performance}

On the $200$-task benchmark, our framework achieves state-of-the-art performance across rendering success, algorithm correctness, aesthetics, and efficiency, as summarized in Figure~\ref{fig:experiment_summary}.
Compared with the strongest end-to-end Manim baseline, our \vta{}-based tool-generation pipeline achieves substantially higher end-to-end success and correctness with lower failure rates; relative to the agentic \texttt{Code2Video} baseline, our RSL-guided renderer matches its rendering success while providing verifiable traces, higher AES scores, and significantly shorter generation time.
Detailed metric values are reported in Tables~\ref{tab:tier2-correctness},~\ref{tab:vta-vs-manim-success}, and~\ref{tab:aes-systems}, with Appendix~\ref{sec:appendix-prompts}, Table~\ref{tab:tier2-rubric-summary} further breaking down the rubric scores by dimension.
Table~\ref{tab:tier2-correctness} shows that VTA+RSL reaches $99.8\%$ algorithm correctness (DeepSeek-V3.1), with $99.2\%$ on average for VTA-only trackers, versus mid-80\% for Manim baselines.

Across all families, tool-making consistently outperforms end-to-end script generation, with larger gains on long-horizon DP and graph/hashtable tasks (Table~\ref{tab:family_success}).
Under our prompting and repair budget, the average LLM cost is $\sim$\$$0.022$ per task (VTA+RSL, $200$ tasks).

Analysis of the generation process shows that the error-guided repair mechanism is critical: the success rate without repair (Pass@$1$) averages $91.0\%$ (e.g., $89.5\%$ for Qwen$3$-$235$B), rising to $99.5\%$ after up to $3$ repair rounds. This demonstrates the efficacy of our compiler-like feedback loop~\cite{madaan2023selfrefineiterativerefinementselffeedback}.

Adding $12$k-token RAG documentation for Manim APIs actually \emph{reduces} the baseline success rate from $82.5\%$ to $79.0\%$, and over $80\%$ of failures remain API misuse or logic drift. This suggests that the bottleneck is task formulation rather than library knowledge; detailed failure statistics are deferred to Appendix~\ref{sec:appendix-failures}.

\subsection{Visual Quality: AES Evaluation}
\label{sec:aes}

We use AES to evaluate videos across five $0$--$20$ dimensions: layout, attractiveness, logic, accuracy \& depth, and consistency.
System-level comparisons are in Table~\ref{tab:aes-systems}; its time and token columns show that our VTA+RSL variant remains efficient while achieving higher AES scores than end-to-end Manim baselines.
Per-model AES scores indicate that model choice has relatively minor impact on aesthetics (Appendix~\ref{sec:appendix-eval}, Table~\ref{tab:aes-models}), while our \vta{}-only pipeline is already competitive with end-to-end Manim baselines.
As an additional end-to-end reference, \texttt{code2video\_manim\_direct} achieves a substantially lower AES total of $74.90$ on the $200$ videos that were successfully evaluated.
The RSL-enhanced variant further improves total AES to $90.17$, mainly by better layout and color choices, without sacrificing correctness.

We also reuse TEA's multimodal evaluation rubric~\cite{ku2025theoremexplainagentvideobasedmultimodalexplanations} as an auxiliary metric, consistent with the AES findings (full TEA results in Appendix~\ref{sec:appendix-tea}).

\subsection{Case Study}
Figure~\ref{fig:case_gallery} shows examples from six algorithm families, demonstrating that our pipeline consistently visualizes core data structures and step-wise execution without occlusion or overlap.
Array/DP examples highlight element updates and 2D table filling via synchronized cell highlighting.
Graph/tree examples co-display structure with auxiliary states (e.g., queues/stacks), while hash-table examples show evolving key--value mappings (including bucket views).
Overall, the main view and auxiliary panels remain spatially separated, making intermediate states easy to follow.

We observe a presentation limitation under high information density: long pseudocode blocks or large tables can force the canvas to rescale elements, reducing legibility (Figure~\ref{fig:case_defect}).
This affects layout/readability rather than semantic correctness, and could be mitigated by adaptive zooming, summarization, or paged/scrollable views.

\section{Conclusion}
\label{sec:conclusion}

We presented a unified framework for algorithm visualization that uses LLMs as \emph{tool makers}. Instead of prompting the model to directly author renderer-level video scripts (e.g., Manim code), the model synthesizes an instrumented Python tracker that emits schema-validated \vtajson{}~$5.0$ traces, which deterministic engines then render into Manim videos, LaTeX figures, and Three.js interactions.
We further introduced \rsl{} to enable controlled, LLM-driven style adaptation without altering execution semantics.

On a $200$-task LeetCode benchmark, our method significantly improves rendering success, algorithm correctness, and layout quality over strong end-to-end baselines.
Beyond algorithm visualization, we believe the tool-maker paradigm can benefit other domains where reliability and controllability are critical, such as multi-tool AI agents and LLM-based data visualization systems, as well as broader UI automation and scientific simulation scenarios.

\section*{Acknowledgements}
This work was supported by the National Natural Science Foundation of China (No.~62576299, No.~U21B2037, No.~U22B2051, No.~U23A20383, No.~62176222, No.~62176223, No.~62176226, No.~62072386, No.~62072387, No.~62072389, No.~62002305, No.~62272401), the Natural Science Foundation of Fujian Province of China (No.~2021J06003, No.~2022J06001), and the Fundamental Research Funds for the Central Universities.

\section*{Limitations}

Our work has several limitations.

First, the benchmark focuses on typical algorithmic tasks with relatively structured states (arrays, graphs, DP tables).
More complex domains (e.g., geometric algorithms with continuous geometry or approximate methods) may require extending \vta{} with richer geometry support or probabilistic annotations.

Second, although we support multiple rendering backends, our empirical evaluation focuses primarily on Manim videos.
Our experiments with TikZ figures and $3$D interactions are more qualitative and limited in scale.
A more systematic user study comparing different backends is left for future work.

Third, we rely on existing open-source LLMs and prompt engineering.
While the tool-maker paradigm is model-agnostic, actual performance and cost trade-offs may vary across LLM families and deployment settings.
Exploring fine-tuned, open-source models for this task is an interesting direction.

Finally, our current evaluation mostly targets technical correctness and visual quality.
We do not yet perform large-scale classroom studies measuring actual learning gains.
Understanding how different visualization styles and levels of detail affect student learning remains an important open question.

\section*{Ethical Considerations}

Our system is designed for educational purposes and does not directly handle sensitive user data.
However, large-scale generation of educational content could impact existing teaching materials and content creators.
We encourage responsible use, including clear attribution when reusing generated materials and avoiding misleading learners with incorrect visualizations.

Our benchmark is constructed from public LeetCode problems and does not include personal or proprietary data.
We release our code under the MIT license and dataset to support reproducibility and further research.
We do not foresee direct negative societal impacts, but as with any automation tool, misuse (e.g., mass-producing low-quality teaching materials without validation) is possible and should be guarded against.
We used AI assistants (e.g., ChatGPT) for language polishing; all content was reviewed and verified by the authors.

\bibliography{custom}

\appendix

\section{Qualitative Gallery}
\label{app:gallery}

\begin{figure*}[t]
    \centering
    \includegraphics[width=\textwidth,height=0.9\textheight,keepaspectratio]{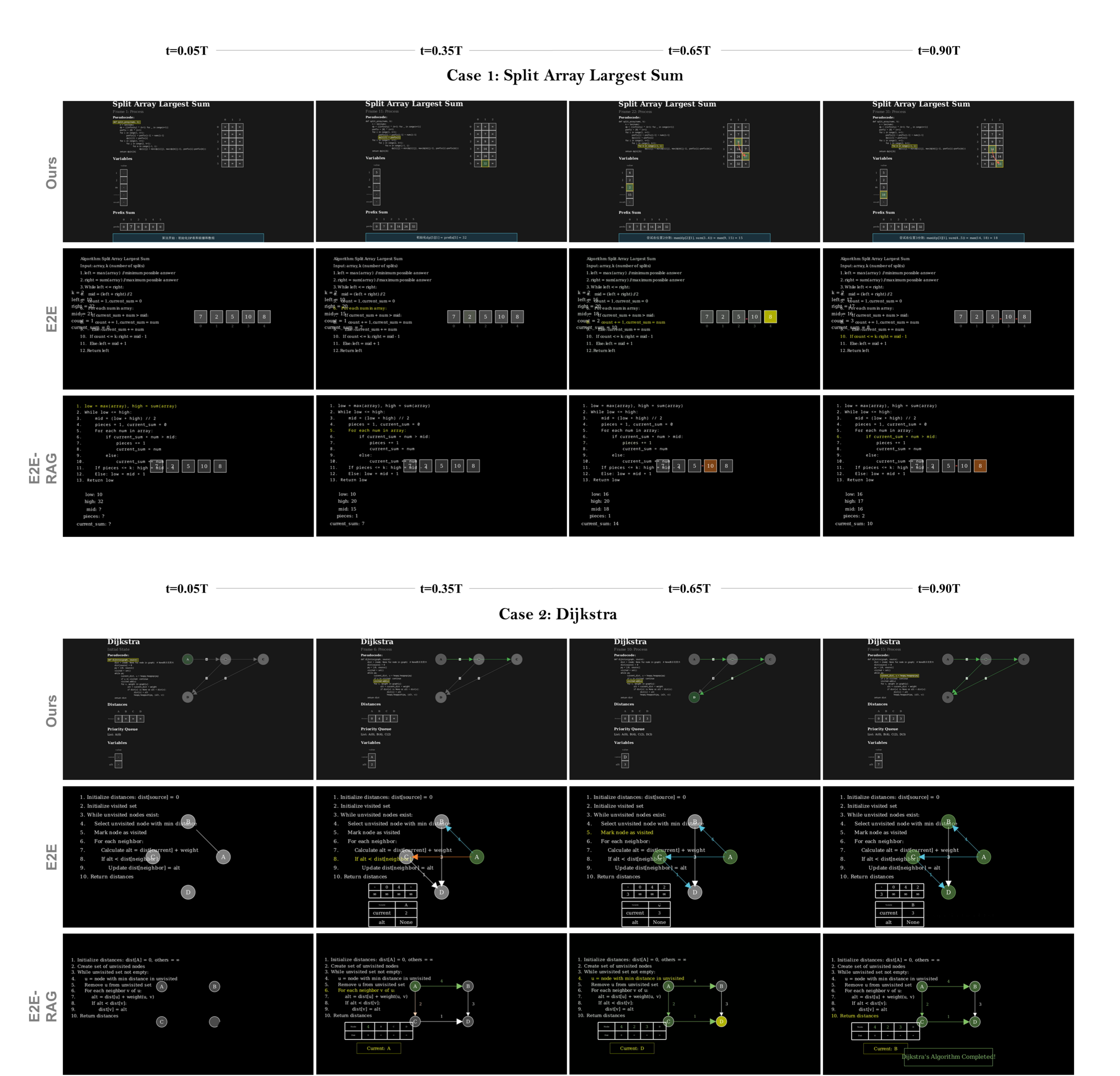}
    \caption{\textbf{Gallery of \ours{} Results and Baselines.} Selected examples from array, graph, dynamic programming, tree, and sorting families, showing representative key frames across methods in a filmstrip layout.}
    \label{fig:gallery}
\end{figure*}

\begin{figure*}[t]
  \centering
  \includegraphics[width=\textwidth]{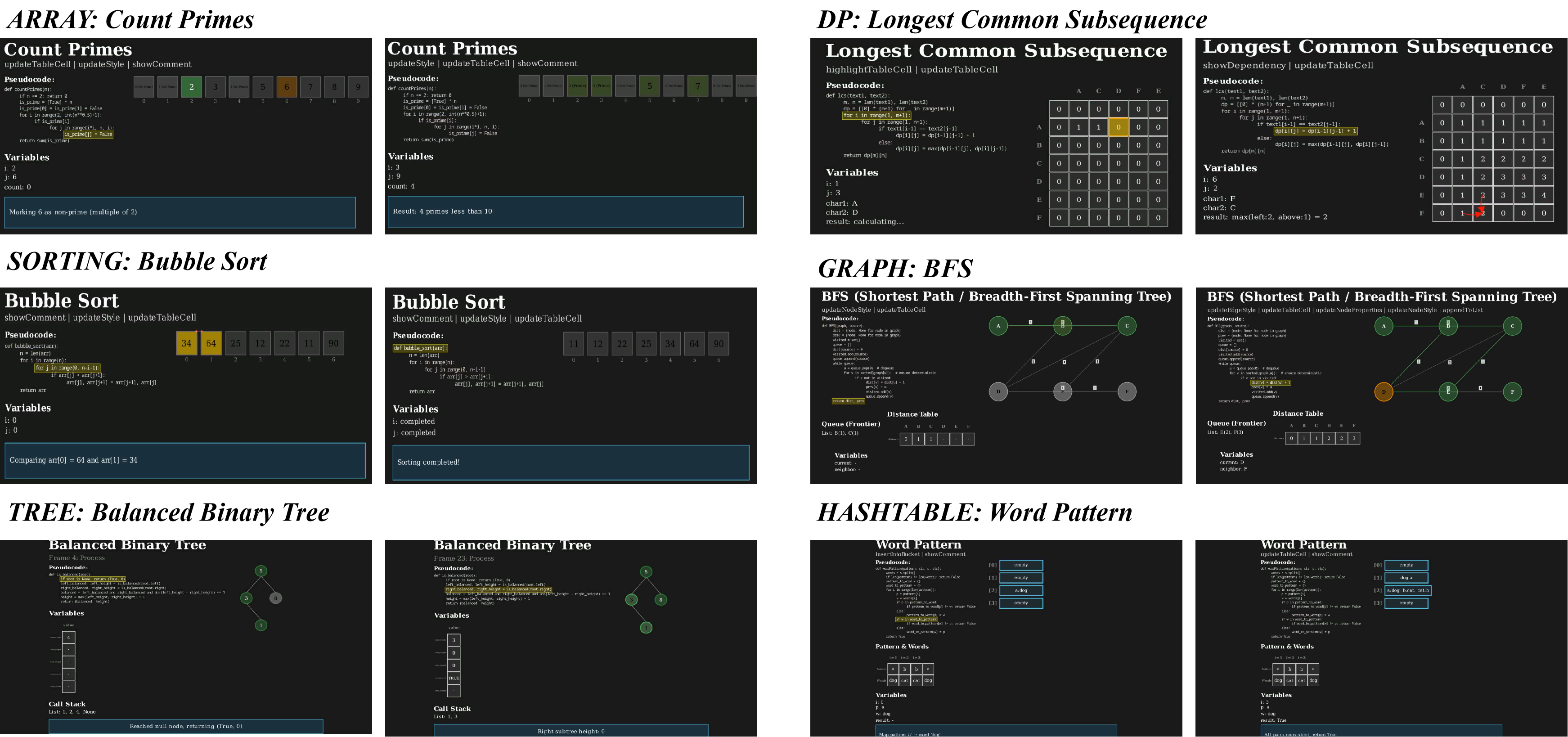}
  \caption{\textbf{Case study: representative visualizations from \ours{}.} Qualitative examples from six algorithm families illustrating that our pipeline can consistently visualize core data structures and step-wise execution without obvious occlusion or overlap. This figure corresponds to the high-quality cases discussed in the case study.}
  \label{fig:case_gallery}
\end{figure*}

\begin{figure*}[t]
  \centering
  \includegraphics[width=\textwidth]{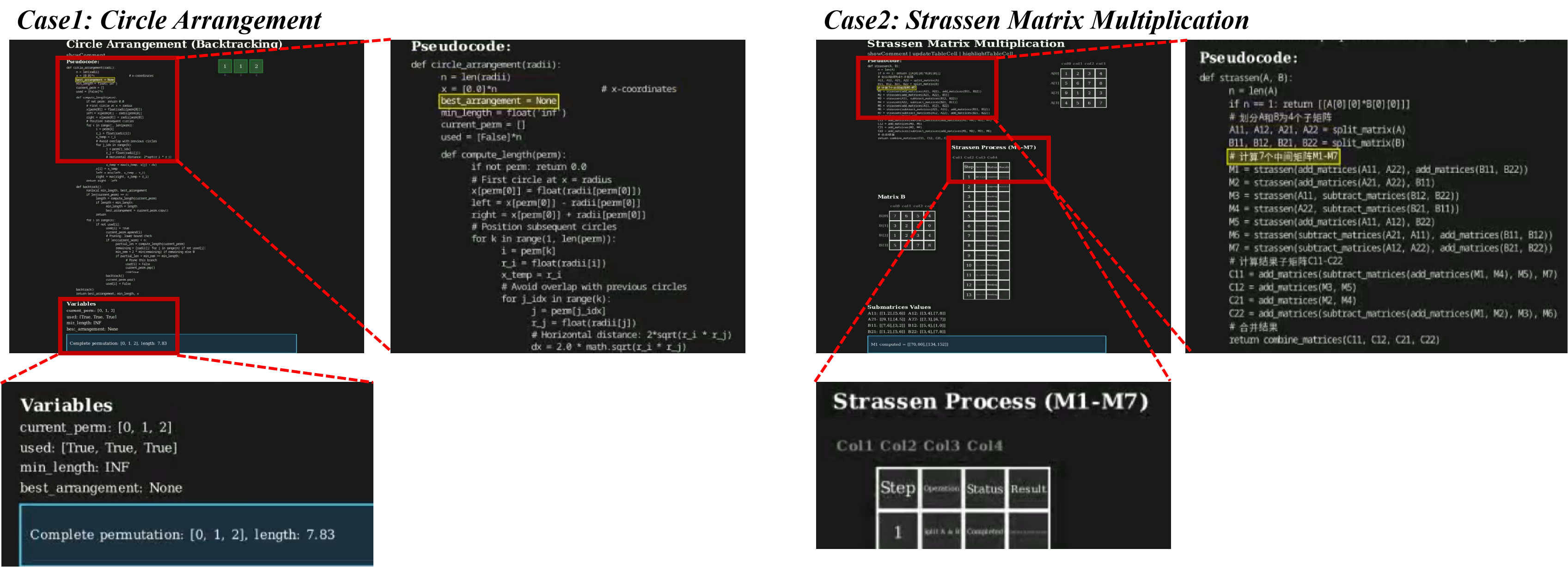}
  \caption{\textbf{Case study: failure modes of \ours{}.} Representative failure cases under high information density, where long pseudocode blocks or large tables cause the canvas to compress visual elements and reduce legibility. This figure corresponds to the limitations discussed in the case study subsection of the main text.}
  \label{fig:case_defect}
\end{figure*}

\begin{figure*}[t]
  \centering
  \includegraphics[width=\textwidth]{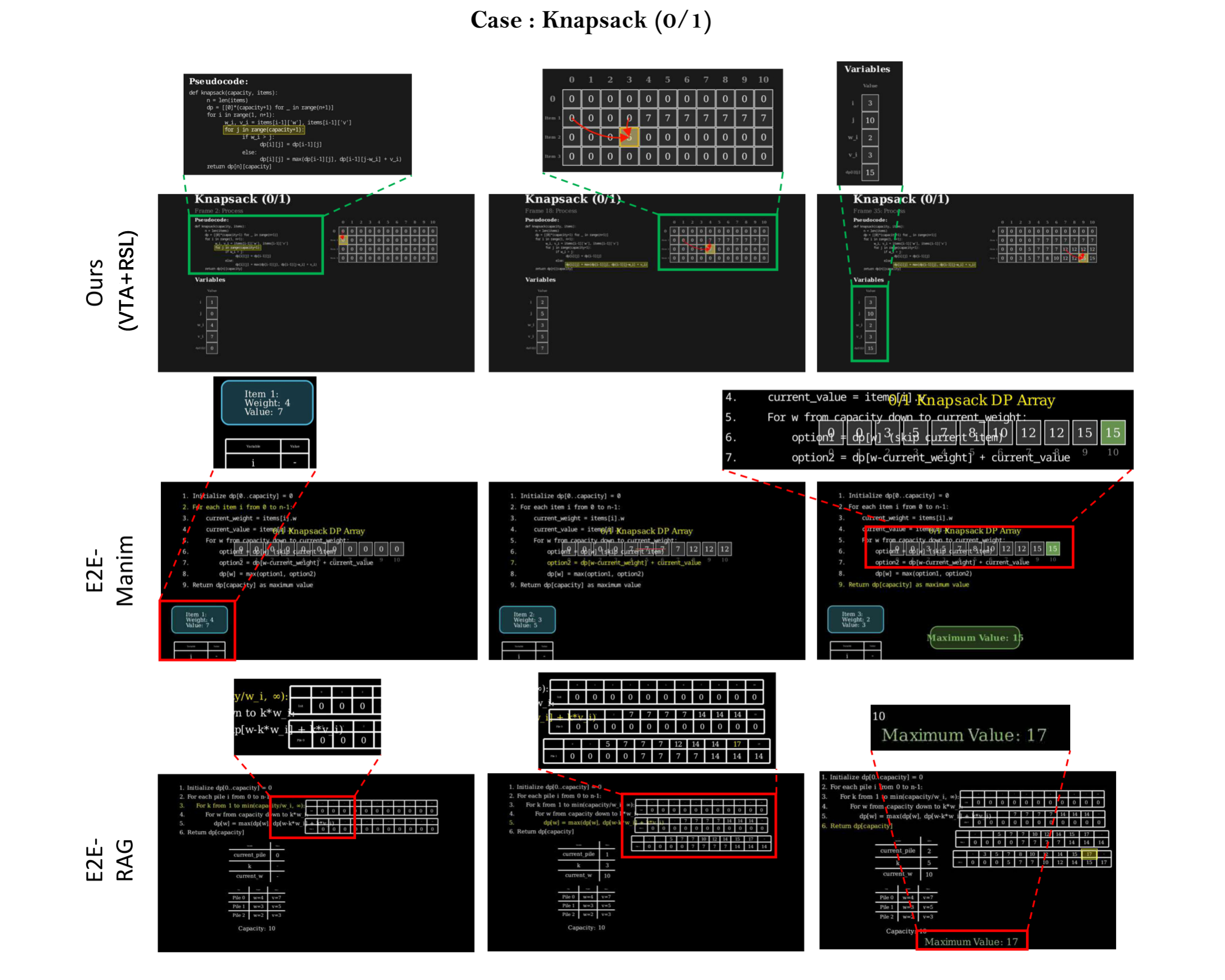}
  \caption{
    \textbf{Qualitative comparison on a complex dynamic-programming task (Knapsack 0/1).}
    Rows show key frames from \textbf{Ours} (Top), \textbf{E2E-Manim} (Middle), and \textbf{E2E-RAG} (Bottom).
    \textbf{(Top)} Our \vta{}-based system generates a clean split-screen layout, with precise variable tracking (green boxes) and synchronized code highlighting.
    \textbf{(Middle)} The end-to-end baseline suffers from \textit{layout clutter}, failing to wrap the long DP array.
    \textbf{(Bottom)} The RAG-enhanced model exhibits severe \textit{spatial hallucinations}, causing elements to overlap and rendering the code unreadable (red boxes).
  }
  \label{fig:failure_case_study}
\end{figure*}

We provide a qualitative gallery of generated visualizations across different algorithm families to demonstrate the versatility of our framework beyond a single running example.
Figure~\ref{fig:gallery} summarizes cross-method comparisons, while Figures~\ref{fig:case_gallery}--\ref{fig:case_defect} provide per-family views of our system's typical visualizations and failure modes used in the case study; Figure~\ref{fig:failure_case_study} further contrasts our method with end-to-end Manim baselines on a challenging dynamic-programming task.

\section{Open-Source Resources}
\label{sec:appendix-resources}

We plan to release:

\begin{itemize}
  \item \textbf{Code}: the full \ours{} system, including \vtajson{} 5.0 specification, validators, renderers, and evaluation scripts.
  \item \textbf{Dataset}: 200 LeetCode tasks in standardized \texttt{example/*.txt} format.
  \item \textbf{Evaluation results}: AES / TEA scores and automatic metrics for all methods.
  \item \textbf{Demo videos}: a curated set of high-quality AV videos across algorithm families.
  \item \textbf{Documentation}: a system technical report and user guide.
\end{itemize}

All resources will be released under the MIT license.

\section{Experimental Environment}

\subsection{Python Dependencies}

\begin{promptblock}
python>=3.9
openai>=1.0.0
manim>=0.18.0
\end{promptblock}

\subsection{System Dependencies}

\begin{promptblock}
# LaTeX (optional, for TikZ rendering)
sudo apt-get install texlive-xetex

# FFmpeg (Manim video encoding)
sudo apt-get install ffmpeg

# Image conversion (optional)
sudo apt-get install pdftocairo
\end{promptblock}

We use a SiliconFlow API key to access all LLM models (DeepSeek-V3, Qwen series, GLM-4.6).

\section{Additional Evaluation Tables}
\label{sec:appendix-eval}

\begin{table*}[t]
  \centering
  \small
  \begin{tabular}{lccccccc}
    \midrule
    Model & \#Eval & Layout & Attract. & Logic & Acc.\&Depth & Consist. & Total \\
    \midrule
    DeepSeek-V3.1 & 200 & 17.37 & 16.72 & 18.08 & \textbf{18.44} & 18.01 & \textbf{88.62} \\
    GLM-4.6 & 200 & 17.36 & \textbf{16.75} & 18.05 & 18.37 & \textbf{18.06} & 88.60 \\
    Qwen3-235B & 199 & 17.36 & 16.54 & 17.90 & 18.29 & 18.03 & 88.13 \\
    \midrule
    Average & 599 & 17.36 & 16.67 & 18.01 & 18.37 & 18.03 & 88.45 \\
    \bottomrule
  \end{tabular}
  \caption{AES evaluation results per model (0--20 per dimension, 0--100 total).}
  \label{tab:aes-models}
\end{table*}

\subsection{Renderer Performance}
\label{sec:appendix-runtime}

Table~\ref{tab:renderer-performance} reports the runtime characteristics of the three renderers on the 200-task benchmark.

\begin{table*}[t]
  \centering
  \small
  \begin{tabular}{lccccc}
    \toprule
    Renderer & \#Traces & Success & Success rate & Avg. time & Output size \\
    \midrule
    Manim video & 200 & 200 & 100\% & 3.03min / task & 2.8MB / video \\
    LaTeX/TikZ & 200 & 200 & 100\% & 1.04min / task & 156KB / frame set \\
    Three.js & 200 & 200 & 100\% & instantaneous & trace.json only \\
    \bottomrule
  \end{tabular}
  \caption{Performance of three renderers on the same 200 \vtajson{} traces.}
  \label{tab:renderer-performance}
\end{table*}

\subsection{TEA Evaluation Details}
\label{sec:appendix-tea}

Table~\ref{tab:tea-models} reports TEA scores per model under the 0--5 rubric.

\begin{table*}[t]
  \centering
  \small
  \begin{tabular}{lccccccc}
    \toprule
    Model & \#Eval & Layout & Attract. & Logic & Acc.\&Depth & Consist. & Total \\
    \midrule
    DeepSeek-V3.1 & 200 & 2.92 & 1.12 & 1.75 & \textbf{4.99} & \textbf{4.93} & \textbf{4.70} \\
    GLM-4.6 & 200 & 2.81 & 1.09 & 1.56 & \textbf{4.99} & \textbf{4.93} & 4.66 \\
    Qwen3-235B & 199 & 2.67 & 1.04 & 1.20 & \textbf{4.99} & \textbf{4.94} & 4.62 \\
    \midrule
    Average & 599 & 2.79 & 1.09 & 1.51 & 4.99 & 4.93 & 4.65 \\
    \bottomrule
  \end{tabular}
  \caption{TEA evaluation results (0--5 scale) for three models.}
  \label{tab:tea-models}
\end{table*}

\section{Dataset Construction Details}\label{sec:appendix-dataset}

\subsection{Data Collection Pipeline}

Our dataset is constructed through a multi-step pipeline (\texttt{batch\_fetch\_all.py}). We first collect 2,530 LeetCode problems via LeetCode's GraphQL endpoint, and then convert them into 3,958 standardized task instances in our system format (each instance corresponds to a concrete input example for a problem):

\begin{enumerate}
  \item \textbf{Batch Fetching}: Retrieve problems from LeetCode's GraphQL endpoint for 6 algorithm families.
  \item \textbf{Manual Reclassification}: Apply corrections for mislabeled problems (e.g., problems tagged as ``Graph'' but actually requiring DP).
  \item \textbf{Format Conversion}: Convert to standardized task specification format, including schema normalization and validation (e.g., graph canonicalization into a nested \texttt{graph} object with explicit \texttt{directed} edges, and basic type/field checks). We export two variants of task files (with and without natural-language problem descriptions) to support different evaluation settings.
  \item \textbf{Stratified Sampling}: From 3,958 candidate task instances, sample 200 tasks using \texttt{create\_small\_dataset.py} with:
  \begin{itemize}
    \item Minimum 15 samples per algorithm family
    \item Remaining samples allocated proportionally
    \item Random seed = 42 for reproducibility
  \end{itemize}
\end{enumerate}

We collect problems via LeetCode's GraphQL endpoint for research/education only, and our released artifacts are intended for the same purpose.
The dataset is a derived research artifact; users should comply with LeetCode's terms of use, and we do not intend it for use outside a research setting.
We avoid redistributing any content beyond what is permitted by the original access conditions.

\subsection{Task Selection Criteria}

\begin{enumerate}
  \item \textbf{Stateful}: Exclude pure mathematical calculations without state changes (e.g., \texttt{Power(x, n)}).
  \item \textbf{Visualizable}: Require data structures with clear geometric representations.
  \item \textbf{Balanced difficulty}: Easy ($35\%$), Medium ($53\%$), Hard ($12\%$).
  \item \textbf{Family diversity}: Array (84), DP (29), Sorting (26), Graph (22), Tree (21), Hashtable (18).
\end{enumerate}

\subsection{Task File Format}\label{sec:task-format}

Each task file (\texttt{example/*.txt}) contains:

\begin{promptblock}
Algorithm Snippet (Course Schedule):

- LeetCode Problem ID: 207
- Difficulty: Medium
- Goal: Generate `graph_tracker.py`
- User Request: Create visualization tracker 
               for "Course Schedule (Graph)"

- Input:
input_data = {
  "graph": {
    "nodes": [{"id": "A", "label": "A"}, ...],
    "edges": [{"from": "A", "to": "B", 
               "weight": 4, "directed": true}, ...]
  },
  "source": "A"
}
\end{promptblock}

\subsection{Quality Control}

All tasks are manually reviewed to ensure:

\begin{itemize}
  \item Valid and representative input data with edge cases.
  \item Correct algorithm family classification after reclassification.
  \item Consistent tagging and visual requirements.
\end{itemize}

\section{\vtajson{} 5.0 Specification}\label{sec:appendix-vtajson}

\vtajson{} 5.0 is the JSON encoding of our Visualization Trace Algebra (\vta{}) for algorithm visualization, designed to be simple for LLMs to generate via in-context learning while expressive enough to cover common AV patterns.

\subsection{JSON Structure}

A \vtajson{} trace has the following top-level structure:

\begin{promptblock}
{
  "vta_version": "5.0",
  "algorithm": {
    "name": "Dijkstra Shortest Path",
    "family": "Graph"
  },
  "initial_frame": {
    "data_schema": { ... },
    "data_state": {
      "type": "graph",
      "structure": {
        "nodes": [{"id": "A", "label": "A",
                   "styleKey": "idle", 
                   "properties": {"distance": 0}}],
        "edges": [{"from": "A", "to": "B",
                   "weight": 4, "styleKey": "normal"}]
      }
    },
    "auxiliary_views": [...],
    "styles": { "elementStyles": {...} },
    "pseudocode": ["1. Initialize distances", ...]
  },
  "deltas": [
    {
      "action_description": "Select node A",
      "code_highlight": 2,
      "operations": [[
        {"op": "updateNodeStyle", 
         "params": {"ids": ["A"], "styleKey": "current"}}
      ]]
    }
  ],
  "required_extensions": ["vta-ext-primitive-graph"]
}
\end{promptblock}

\subsection{Supported Operations}\label{sec:vta-operations}

\vta{} defines approximately 30 atomic operations organized by view type, which are serialized in \vtajson{} traces as \texttt{op} codes:

\begin{table}[ht]
\centering
\small
\begin{tabular}{ll}
\toprule
\textbf{View Type} & \textbf{Core Operations} \\
\midrule
Array & \texttt{updateStyle}, \texttt{moveElements}, \\
      & \texttt{shiftElements}, \texttt{updateValues} \\
\midrule
Graph & \texttt{updateNodeStyle}, \texttt{updateNodeProperties}, \\
      & \texttt{updateEdgeStyle}, \texttt{addNode}, \texttt{removeNode} \\
\midrule
Tree & \texttt{addChild}, \texttt{reparent}, \texttt{rotate} \\
\midrule
Hashtable & \texttt{insertIntoBucket}, \texttt{rehash}, \\
          & \texttt{highlightCollision} \\
\midrule
Table (DP) & \texttt{updateTableCell}, \texttt{highlightTableCell}, \\
           & \texttt{showDependency} \\
\midrule
Generic & \texttt{showComment}, \texttt{hideComment}, \\
        & \texttt{appendToList}, \texttt{popFromList} \\
\bottomrule
\end{tabular}
\caption{\vta{}/\vtajson{} 5.0 operation categories by view type.}
\label{tab:vta-operations}
\end{table}

\subsection{Delta Semantics}

\vtajson{} uses \emph{deltas} rather than absolute frames. Each delta describes how to transform the previous frame into the next one, making it natural to map onto animations (e.g., smooth movement, fade-in/fade-out) and to align trace steps with algorithm operations. The \texttt{operations} field is a 2D array, where the inner arrays group logically simultaneous operations.

\subsection{Schema Validation}

A validator enforces \vtajson{} 5.0 invariants:
\begin{itemize}
  \item \texttt{vta\_version} must be string \texttt{"5.0"} (not numeric)
  \item \texttt{operations} must be 2D arrays (\texttt{[[...]]})
  \item No \texttt{Infinity} values (use \texttt{null} for undefined)
  \item Graph edge endpoints must reference existing nodes
  \item \texttt{code\_highlight} must be integer or integer array
\end{itemize}

\section{\rsl{} Specification}\label{sec:appendix-rsl}

\rsl{} (Rendering Style Language) is a declarative DSL that controls \emph{how} \vtajson{} traces are rendered. It is implemented in \texttt{manim/test\_v2/} with three components: \texttt{rsl\_schema.json} (JSON Schema), \texttt{rsl\_generator.py} (LLM-based generation), and \texttt{rsl\_interpreter.py} (conversion to render config).

\subsection{Schema Structure}

An \rsl{} configuration contains five top-level fields:

\begin{promptblock}
{"meta": {"rsl_version": "0.1"},
 "theme": {"background": "#1A1A1A",
   "text": "#FFFFFF", "primary": "#3498DB"},
 "timeline": {"transition": 0.5, "pause": 0.3},
 "layout": {"main": {
   "type": "force_directed",  // or grid/matrix
   "params": {"node_spacing": 2.0}}},
 "rules": [{"when": {"op": "updateNodeStyle"},
   "do": {"animation": {"variant": "pulse"}}}]}
\end{promptblock}

\subsection{Layout Types}

Supported layout types: \texttt{force\_directed}, \texttt{hierarchical}, \texttt{circular}, \texttt{grid}, \texttt{matrix}, \texttt{horizontal\_array}. Layout parameters include \texttt{node\_spacing} (1.0--10.0), \texttt{edge\_curve} (-1.0--1.0), \texttt{cell\_size} (0.3--2.0).

\subsection{LLM-Driven Generation}

The \texttt{rsl\_generator.py} extracts trace features (algorithm family, data type, scale, operations used) and prompts an LLM to generate an \rsl{} config. The output is validated against the JSON Schema and semantic checks (e.g., only allowed \vta{} operation names in \texttt{rules[].when.op}).

\subsection{Safety Constraints}

\begin{itemize}
  \item Numeric params bounded: \texttt{transition} $\in [0.1, 2.0]$, \texttt{pause} $\in [0, 1.0]$
  \item Layout types constrained to predefined enums
  \item Animation variants: \texttt{pulse}, \texttt{glow}, \texttt{shake}, \texttt{fade}, \texttt{morph}
  \item Invalid configs fall back to defaults
\end{itemize}

\section{Formal Properties of \vta{}}\label{sec:appendix-vta}

\subsection{Monoid Structure of Primitive Operations}

\paragraph{Theorem A.1 (VTA monoid of primitive operations).}
Let $Op$ be the finite set of primitive visual operation symbols defined in our \vtajson{}~5.0 specification (Appendix~\ref{sec:appendix-vtajson}; e.g., \texttt{updateStyle}, \texttt{updateNodeStyle}, \texttt{updateTableCell}).
Let $M := Op^*$ be the set of all finite sequences (including the empty sequence) over $Op$.
Define a binary operation $* : M \times M \to M$ by sequence concatenation: for $a,b \in M$, $a * b$ is the sequence obtained by appending $b$ to $a$.
Let $\epsilon \in M$ denote the empty sequence.
Then $(M, *, \epsilon)$ is a monoid.

\paragraph{Proof.}
We first fix notation.
Each element of $Op$ denotes a primitive visual operation such as updating a table cell or changing a node style.
An element of $M = Op^*$ is therefore a finite sequence
$w = (o_1,\dots,o_n)$ of such primitive operations; the empty sequence is denoted by $\epsilon$.

\emph{Closure.}
By construction, every element of $M$ is a finite sequence of symbols from $Op$.
If $a,b \in M$, then $a * b$ is obtained by concatenating two finite sequences over $Op$.
The result is again a finite sequence over $Op$, so $a * b \in M$.
Therefore the binary operation $* : M \times M \to M$ is closed on $M$.

\emph{Associativity.}
Concatenation of finite sequences is associative: for any $a,b,c \in M$,
first concatenating $a$ and $b$ and then concatenating the result with $c$ yields exactly the same sequence as first concatenating $b$ and $c$ and then concatenating the result with $a$.
Formally,
\[
  (a * b) * c = a * (b * c)
  \qquad\text{for all } a,b,c \in M.
\]
This is the standard associativity property of free monoids over a generating set $Op$.

\emph{Identity element.}
Let $\epsilon \in M$ be the empty sequence.
For any $a \in M$, concatenating $\epsilon$ on the left or on the right leaves $a$ unchanged:
\[
  \epsilon * a = a,
  \qquad
  a * \epsilon = a.
\]
Thus $\epsilon$ is a two-sided identity element for $(M, *)$.

Combining closure, associativity, and the existence of an identity element,
we conclude that $(M, *, \epsilon)$ is a monoid.
\hfill$\square$

\paragraph{Remark.}
In the main text, we equip the visual state space $S$ with a right action of this monoid.
Each primitive operation $o \in Op$ is interpreted as a (partial) state transformer
$\llbracket o \rrbracket : S \rightharpoonup S$,
and a sequence $w = (o_1,\dots,o_n) \in M$ acts on a state $s \in S$ via the composite
$\llbracket o_n \rrbracket \circ \dots \circ \llbracket o_1 \rrbracket (s)$.
This satisfies the usual action law $s \cdot (a * b) = (s \cdot a) \cdot b$,
so \vta{} can be viewed as a typed visual state space equipped with a monoid action of primitive operations.

\section{Prompt Engineering}\label{sec:appendix-prompts}

Our prompts are composed of modular components. Due to space limits, we present the structural skeleton and key instructions below. Full prompts are available in the code repository.

\subsection{Tracker Generation Prompt}

  The tracker generation prompt (\texttt{vta\_unified\_v2.txt}, $\sim$43KB) contains six modules:

\begin{table}[ht]
\centering
\small
\begin{tabular}{p{2.2cm}p{5.0cm}}
\toprule
\textbf{Module} & \textbf{Purpose} \\
\midrule
Core Philosophy & Three-version self-verification process \\
Algorithm First & Correctness as highest priority \\
13 Hard Rules & Non-negotiable \vtajson{} constraints \\
Code Structure & Required ordering of code sections \\
Family Guidelines & Graph/DP/Sorting specific rules \\
\vtajson{} 5.0 Spec & Complete operation definitions \\
\bottomrule
\end{tabular}
  \caption{Tracker prompt module overview.}
  \end{table}
 
\begin{tcolorbox}[
    colback=gray!7,
    colframe=gray!50,
    title=\textbf{System Prompt: Tracker Generation ($\mathcal{P}_{\text{tracker}}$)},
    fonttitle=\bfseries,
    arc=0mm, boxrule=0.4pt,
    breakable
]
\small
\textbf{Role:} You are an algorithm visualization expert generating Python trackers that emit \vtajson{}~5.0 traces.

\textbf{Three-Version Process:}
1.~\textbf{Version 1 (Draft)}: write an initial tracker focusing on algorithm logic.\\
2.~\textbf{Self-Check}: verify all hard rules, \vtajson{} field naming, and input/output alignment.\\
3.~\textbf{Version 2 (Corrected)}: fix every issue found in self-check.\\
4.~\textbf{Final Verification}: mentally simulate the first few iterations and check that data-structure updates and code highlights are correct.\\
5.~\textbf{Version 3 (Final Submission)}: the only version we keep and execute.

\textbf{Hard Constraints:}
We define a set of non-negotiable constraints covering \vtajson{} structural invariants (version strings, array dimensions, field naming), data type restrictions, and code quality requirements. Representative examples include ensuring \texttt{operations} is always a 2D array and that render operations never directly modify Python state. The complete constraint set is available in our codebase.

\textbf{Algorithm-Family Guidelines:}
We provide algorithm-family-specific guidelines for graph algorithms (e.g., frontier management, deterministic traversal order), dynamic programming (e.g., table-based state representation), and sorting algorithms (e.g., element comparison and swap visualization). These guidelines ensure consistent and pedagogically sound visualizations across algorithm families.

\textbf{\vtajson{}~5.0 Specification:}
\textit{[The full JSON schema and $\sim$30 operation definitions are provided in the repository under \texttt{vta\_specification/} and omitted here for brevity.]}
\end{tcolorbox}

\subsection{Generated Tracker Example}

Below is a simplified excerpt from a generated tracker for ``Count Primes'' (Sieve of Eratosthenes), illustrating the structure of LLM-generated code and how a tracker emits a \vtajson{} trace:

\begin{promptblock}
import json

input_data = {"array": [1, 2, 3, 4, 5]}

def main():
    n = len(input_data["array"])
    is_prime = [True] * (n + 1)
    is_prime[0] = is_prime[1] = False
    
    trace = {
        "vta_version": "5.0",
        "algorithm": {"name": "Count Primes",
                      "family": "Sieve of Eratosthenes"},
        "required_extensions": ["vta-ext-primitive-array"],
        "initial_frame": {
            "data_state": {"type": "array",
                "structure": [{"index": i, "value": v, 
                    "state": "idle"} 
                    for i, v in enumerate(input_data["array"])]},
            "pseudocode": ["1. Initialize sieve", ...],
            "styles": {"elementStyles": {...}}
        },
        "deltas": []
    }
    
    # Algorithm execution with VTA-JSON operations
    for i in range(2, int(n**0.5) + 1):
        if is_prime[i]:
            # Emit VTA-JSON operation before state change
            trace["deltas"].append({
                "code_highlight": 5,
                "operations": [[{"op": "updateStyle",
                    "params": {"indices": [i], 
                               "styleKey": "current"}}]]
            })
            for j in range(i*i, n+1, i):
                is_prime[j] = False  # Python state first
                # Then emit render operation
                trace["deltas"].append({...})
    
    with open("trace.json", "w") as f:
        json.dump(trace, f)

if __name__ == "__main__":
    main()
\end{promptblock}

The key pattern is: \textbf{update Python state first, then emit VTA-JSON render operations}. This ensures the trace accurately reflects algorithm execution while keeping the tracker code readable and teachable.

\subsection{RSL Generation Prompt (\texorpdfstring{$\mathcal{P}_{\text{rsl}}$}{P\_rsl})}

\begin{tcolorbox}[
    colback=gray!5,
    colframe=gray!50,
    title=\textbf{System Prompt: RSL Generation ($\mathcal{P}_{\text{rsl}}$)},
    fonttitle=\bfseries,
    arc=0mm, boxrule=0.4pt,
    breakable
]
\small
\textbf{Role:} You are a rendering-style expert. Output only valid RSL JSON.

\textbf{VTA-JSON Context (Read-Only):}
Algorithm name and family, data type and scale, number of frames, and the list of VTA-JSON operations used are summarized and passed as context; the trace itself must not be modified.

\textbf{Design Goals:}
Improve layout clarity, visual contrast, and animation rhythm while respecting the trace semantics. In particular, \texttt{rules[].when.op} must use actual \vta{} operation names (e.g., \texttt{updateNodeStyle}, \texttt{updateTableCell}), not semantic labels like ``visit\_node''.

\textbf{Schema Snippet:}
The prompt inlines the RSL JSON Schema (meta/theme/timeline/layout/rules/annotations) so that the model can validate field names and ranges.\newline
\textit{[We omit the full schema here; see \texttt{manim/test\_v2/rsl\_schema.json} for details.]}
\end{tcolorbox}

\subsection{Error-Guided Repair Prompt (\texorpdfstring{$\mathcal{P}_{\text{repair}}$}{P\_repair})}

\begin{tcolorbox}[
    colback=gray!5,
    colframe=gray!50,
    title=\textbf{System Prompt: Error-Guided Repair ($\mathcal{P}_{\text{repair}}$)},
    fonttitle=\bfseries,
    arc=0mm, boxrule=0.4pt,
    breakable
]
\small
\textbf{Context:} If executing the tracker or validating the \vtajson{} trace fails, we summarize the error and feed it back to the model.

\textbf{Error Block:}
\begin{promptblock}
[Previous Error]
{error_type}: {error_message}
Location: line {line_number}, variable {var_name}
\end{promptblock}

\textbf{Repair Instructions (Abstracted):}
1.~Analyze the error type, location, and offending variable.\\
2.~Apply targeted fixes guided by the summarized error (e.g., resolving schema violations or type mismatches) while preserving the intended algorithmic behavior.\\
3.~Keep unrelated code intact and focus edits on the minimal changes needed to pass validation.\\
4.~Return a complete, self-contained Python file; do not output patches.
\end{tcolorbox}

\subsection{Algorithm Correctness Evaluator (\texorpdfstring{$\mathcal{P}_{\text{tier2}}$}{P\_tier2})}

\begin{tcolorbox}[
    colback=gray!5,
    colframe=gray!50,
    title=\textbf{System Prompt: Algorithm Correctness Evaluator ($\mathcal{P}_{\text{tier2}}$)},
    fonttitle=\bfseries,
    arc=0mm, boxrule=0.4pt,
    breakable
]
\small
\textbf{Role:} You are an algorithm expert reviewing a Python tracker for a single visualization example.

\textbf{Inputs:}
Algorithm name and family, an optional reference pseudocode snippet (if available), and the full \texttt{tracker.py} source including VTA-JSON-specific trace-generation logic.

\textbf{Rubric (100 points total):}
\begin{itemize}
  \item 50 pts \emph{Algorithm logic correctness}: coverage of key steps and execution order for the given example.
  \item 30 pts \emph{VTA-JSON compliance}: required fields, operation names, data types, and extension declarations.
  \item 10 pts \emph{Result presentation}: how final outputs and important variables are surfaced in the trace.
  \item 10 pts \emph{Code quality}: naming, comments on key steps, and overall structure of the trace-generation code.
\end{itemize}

\textbf{Evaluation principles:}
Focus on whether the tracker correctly implements the intended algorithm for the visualization input rather than speculating about unseen corner cases; distinguish semantic errors (wrong results, missing steps) from stylistic differences (e.g., using \texttt{<=} vs. \texttt{<}); when uncertain about a critical branch, err on the conservative side and record the source of uncertainty in the issue list instead of assigning a near-perfect score.

\textbf{Output format:}
The model must return a strict JSON object with per-dimension scores, a 0--50 \texttt{algorithm\_logic} subscore (later normalized to the reported 0--100 ``algorithm-correctness'' metric), a list of issues (\texttt{severity}/\texttt{category}/\texttt{description}), a list of strengths, and a short overall assessment.
\end{tcolorbox}

We apply $\mathcal{P}_{\text{tier2}}$ to all 200 LeetCode tasks for each tracker generator (DeepSeek-V3.1, Qwen3-235B, GLM-4.6). The static evaluator itself is always DeepSeek-V3.1, acting purely as a code reviewer. On the DeepSeek-generated trackers, this setup consumes on average roughly 5.1k prompt tokens and 0.3k completion tokens per tracker (about 5.4k tokens in total), providing a scalable yet fine-grained view of code-level correctness.

\begin{table}[ht]
  \centering
  \small
  \resizebox{\columnwidth}{!}{
  \begin{tabular}{lcccc}
    \toprule
    Model & Alg. logic & VTA compl. & Return disp. & Code qual. \\
    \midrule
    DeepSeek-V3.1 & 49.9 & 28.5 & 9.9 & 8.8 \\
    Qwen3-235B    & 49.4 & 28.0 & 8.9 & 8.7 \\
    GLM-4.6       & 49.5 & 28.5 & 9.8 & 8.8 \\
    \bottomrule
  \end{tabular}}
  \caption{Average tier-2 rubric scores (0--50 for algorithm logic, 0--30 for VTA-JSON compliance, 0--10 for return presentation and code quality) over 200 tasks per generator model.}
  \label{tab:tier2-rubric-summary}
\end{table}

\paragraph{Per-family end-to-end success.}
Table~\ref{tab:family_success} reports end-to-end success rates by algorithm family and the absolute improvement of our VTA+RSL pipeline over the \texttt{manim\_direct} baseline.

\begin{table*}[t]
\centering
\small
\setlength{\tabcolsep}{3pt}
\renewcommand{\arraystretch}{0.95}
\begin{tabular}{lccccccc}
\toprule
 & Sort & Array & DP & Tree & Graph & Hash & Avg. \\
\midrule
Ours (VTA+RSL)  & 100.0 & 100.0 & 100.0 & 100.0 & 100.0 & 100.0 & 100.0 \\
Manim\_direct   & 88.5  & 89.3  & 58.6  & 95.2  & 72.7  & 77.8  & 80.4 \\
\midrule
Gap ($\uparrow$) & +11.5 & +10.7 & +41.4 & +4.8  & +27.3 & +22.2 & +19.6 \\
\bottomrule
\end{tabular}
\caption{End-to-end success rate (\%) by algorithm family on ALGOGEN-Bench. \textbf{Gap} reports absolute improvements of Ours over Manim\_direct.}
\label{tab:family_success}
\end{table*}

\section{Failure Analysis of End-to-End Baselines}\label{sec:appendix-failures}

\subsection{Breakdown of Failures}

Analysis of 35 failed cases from \texttt{manim\_direct}:

\begin{table}[ht]
\centering
\small
\begin{tabular}{lcc}
\toprule
\textbf{Error Type} & \textbf{Count} & \textbf{Percentage} \\
\midrule
API parameter errors & 18 & 51.4\% \\
Non-existent attributes & 7 & 20.0\% \\
Class confusion & 5 & 14.3\% \\
Rendering timeout & 5 & 14.3\% \\
\bottomrule
\end{tabular}
\caption{Distribution of failure types in end-to-end Manim generation.}
\label{tab:failure-types}
\end{table}

\subsection{Representative Failure Cases}

\paragraph{Case 1: API Parameter Error.}
\begin{promptblock}
# Generated (incorrect)
self.pseudocode[line].set_background_stroke(BLACK, 3)

# Correct usage
self.pseudocode[line].set_background_stroke(
    color=BLACK, width=3)
\end{promptblock}
The LLM used positional arguments, but Manim v0.18 requires keyword arguments. Despite correct documentation in the prompt, the model relied on pre-training patterns from older Manim versions.

\paragraph{Case 2: Non-existent Attribute.}
\begin{promptblock}
# Generated (incorrect)
num_nodes = self.vars_table.shape[0]

# Error: 'Table' has no attribute 'shape'
# Correct: len(self.vars_table.get_rows())
\end{promptblock}
The LLM incorrectly transferred \texttt{numpy} array patterns to Manim \texttt{Table} objects---a ``negative transfer'' from similar libraries.

\paragraph{Case 3: Rendering Timeout.}
For DP algorithms, the LLM generated fine-grained animations for each cell update (1000+ \texttt{FadeIn} calls), exceeding 20-minute timeout. The model cannot estimate animation computational cost.

\subsection{Why RAG Does Not Help}

RAG with 12K-token Manim documentation \emph{reduces} success rate from $82.5\%$ to $79.0\%$. Analysis of 42 RAG failures:

\begin{itemize}
  \item $60\%$ are algorithm logic errors (RAG provides API knowledge, not algorithmic reasoning)
  \item $40\%$ are API knowledge issues (even with correct docs, LLM relies on pre-training memory)
  \item Information overload: 12K tokens dilute attention on critical constraints
\end{itemize}

This demonstrates that the bottleneck is \emph{task architecture}, not knowledge availability. Our \vta{}/\vtajson{} pipeline achieves near-perfect success by simplifying the task structure rather than merely augmenting knowledge.

\section{Baseline Implementation Details}\label{sec:appendix-baseline-prompts}

To ensure a fair comparison, we design comprehensive system prompts for the end-to-end baselines: \texttt{manim\_direct\_novta}, \texttt{manim\_direct}, and \texttt{manim\_direct\_rag}. Despite detailed API guidelines and layout instructions, these baselines still struggle with logical consistency and spatial layout, as summarized in Table~\ref{tab:vta-vs-manim-success}.

\subsection{Prompt for \texttt{manim\_direct\_novta}}

This prompt allows the model to freely design Manim visualizations while enforcing basic correctness and usability constraints, without relying on our structured \vta{}/\vtajson{} IR.

\begin{tcolorbox}[
    colback=gray!7,
    colframe=gray!50,
    title=\textbf{System Prompt: Free Manim Generation ($\mathcal{P}_{\text{free}}$)},
    fonttitle=\bfseries,
    arc=0mm, boxrule=0.4pt,
    breakable
]
\small
\textbf{Role:} You are a Manim author with full creative freedom to design the ``best'' visualization for the given algorithm.

\textbf{Core Principle:} Algorithmic correctness is the top priority.

\textbf{Strict Prohibitions:}
1. Do not ask the user any questions or request more information.
2. Do not output partial code or natural-language explanations.

\textbf{Hard Rules (Abstracted):}
1. Code must run with \texttt{manim -pqh scene.py AlgorithmScene}.
2. Class name must be \texttt{AlgorithmScene(Scene)} with a complete \texttt{construct()}.
3. Pseudocode must be shown somewhere on screen and cover key steps.
4. No Chinese variable names.
5. Graph algorithms must sort adjacency lists to ensure determinism.
6. Use \texttt{VGroup} to manage objects; remove temporary objects to avoid accumulation.
7. Animation run time in each step should be within $[0.3, 0.8]$ seconds.
8. Maintain readable text and sufficient color contrast; elements must stay within the canvas.

\textbf{Manim Cheat Sheet (Provided in Context):}
\textit{[Dozens of examples for text, geometric primitives, tables, layout helpers (\texttt{next\_to}, \texttt{arrange}), and common animations (\texttt{FadeIn}, \texttt{Create}, \texttt{.animate.set\_color}, etc.) are provided here. We omit $\sim$100+ lines of API examples for brevity.]}

\textbf{Recommended Code Skeleton:}
\begin{promptblock}
from manim import *
import numpy as np

input_data = {...}

class AlgorithmScene(Scene):
    def construct(self):
        self.show_title()
        self.create_pseudocode()
        self.create_visualization()
        self.run_algorithm()
        self.show_result()

    # Helper methods are free-form but encouraged.
\end{promptblock}
\end{tcolorbox}

\subsection{Prompt for \texttt{manim\_direct}}

The \texttt{manim\_direct} baseline uses a more structured prompt that mirrors the three-version self-verification idea, but still generates Manim Scenes directly without using \vta{}/\vtajson{}.

\begin{tcolorbox}[
    colback=gray!10,
    colframe=gray!60,
    title=\textbf{System Prompt: Structured End-to-End Generation ($\mathcal{P}_{\text{direct}}$)},
    fonttitle=\bfseries,
    arc=0mm, boxrule=0.4pt,
    breakable
]
\small
\textbf{Role:} You are an expert Manim developer. Write a complete Python script using Manim Community Edition to visualize the given algorithm.

\textbf{Three-Version Process:}
1. \textbf{Version 1 (Draft)}: Produce an initial Manim Scene.
2. \textbf{Self-Check}: Inspect algorithm logic, layout, and API usage; list problems explicitly.
3. \textbf{Version 2 (Corrected)}: Fix all issues found in self-check.
4. \textbf{Final Verification}: Re-check algorithm and Manim APIs.
5. \textbf{Version 3 (Final Submission)}: Polished code that will be executed.

\textbf{Key Constraints (Abstracted):}
\begin{itemize}
  \item Code must run with \texttt{manim -pql scene.py AlgorithmScene}.
  \item Include a detailed pseudocode panel; highlighting must stay synchronized with algorithm steps.
  \item Sort graph adjacency lists for reproducible traversal order.
  \item Use \texttt{VGroup}, \texttt{arrange}, and \texttt{next\_to} to avoid overlaps and keep all elements within the frame.
  \item Use reasonable animation timing (0.3--0.8s), readable fonts, and high-contrast colors.
\end{itemize}

\textbf{Layout Specification (Right-vs-Left Panels):}
- Left panel: stacked pseudocode, variable table, distance table, and frontier queue.
- Right panel: main data view (array visualization, graph, or DP table).

\textbf{Code Skeleton:}
\begin{promptblock}
from manim import *

class AlgorithmScene(Scene):
    def construct(self):
        self.setup_layout()
        self.create_pseudocode()
        self.create_data_view()
        self.create_auxiliary_views()
        self.run_algorithm()
        self.show_result()
\end{promptblock}
\end{tcolorbox}

\subsection{Prompt for \texttt{manim\_direct\_rag}}

The RAG-enhanced baseline uses the same core prompt as \texttt{manim\_direct}, but prepends retrieved Manim API documentation.

\begin{tcolorbox}[
    colback=gray!5,
    colframe=gray!60,
    title=\textbf{System Prompt: RAG-Enhanced Generation},
    fonttitle=\bfseries,
    arc=0mm, boxrule=0.4pt,
    breakable
]
\small
\textbf{Role:} You are an expert Manim developer.

\textbf{Retrieved Context (Top-$K$ API Docs):}
\textit{[Before the main instruction, we insert a large block of Manim API snippets retrieved from \texttt{exp/manim\_direct\_rag/manim\_api\_knowledge.json}, including class definitions (e.g., \texttt{Graph}, \texttt{Table}), typical method usages, and layout/animation examples. The total size of this context is roughly 12K tokens; we omit the $\sim$3000-word documentation here for brevity.]}

\textbf{Instruction (Simplified):}
Using only the documentation and guidelines above, write a complete Manim script that visualizes the given algorithm description, following the same layout and correctness constraints as \texttt{manim\_direct}.
\end{tcolorbox}

\section{Statements}

\subsection{Conflict of Interest}

The authors declare no conflict of interest.

\subsection{Data and Code Availability}

All code, datasets, and the \vtajson{} specification used in this paper will be made publicly available on GitHub under the MIT license upon publication.

\end{document}